\documentclass{article}

\usepackage[utf8]{inputenc} 
\usepackage[T1]{fontenc}    
\usepackage{xcolor}
\definecolor{darkblue}{rgb}{0,0.08,0.45}

\usepackage{url}            
\usepackage{booktabs}       
\usepackage{amsfonts}       
\usepackage{nicefrac}       
\usepackage{microtype}      

\usepackage[preprint]{neurips_2020}

\usepackage{hyperref}
\hypersetup{ %
    pdfkeywords={},
    pdfborder=0 0 0,
    pdfpagemode=UseNone,
    colorlinks=true,
    linkcolor=darkblue,
    citecolor=darkblue,
    filecolor=darkblue,
    urlcolor=darkblue,
    pdfview=FitH}

\usepackage{graphicx}
\usepackage{subfigure}
\usepackage{soul}

\usepackage{amsmath}
\usepackage{mathtools}
\usepackage{amsthm}
\usepackage[normalem]{ulem}

\usepackage[capitalize,noabbrev]{cleveref}

\theoremstyle{plain}

\theoremstyle{definition}

\theoremstyle{remark}

\usepackage{xcolor}
\usepackage{soul}
\usepackage{colortbl}
\definecolor{adaptation}{HTML}{FCCDE5}
\definecolor{objective}{HTML}{FFC2BA}
\definecolor{architecture}{HTML}{FFFFB3}
\definecolor{evaluation}{HTML}{B9DEFF}
\definecolor{finetuning}{HTML}{D7F3E7}
\definecolor{neutral}{HTML}{CFCFCF}

\DeclareRobustCommand{\adaptation}[1]{\sethlcolor{adaptation}{\textbf{\hl{~#1~}}}}
\DeclareRobustCommand{\objective}[1]{\sethlcolor{objective}{\textbf{\hl{~#1~}}}}
\DeclareRobustCommand{\architecture}[1]{\sethlcolor{architecture}{\textbf{\hl{~#1~}}}}
\DeclareRobustCommand{\evaluation}[1]{\sethlcolor{evaluation}{\textbf{\hl{~#1~}}}}
\DeclareRobustCommand{\neutral}[1]{\sethlcolor{neutral}{\textbf{\hl{~#1~}}}}
\DeclareRobustCommand{\finetuning}[1]{\sethlcolor{finetuning}{\textbf{\hl{~#1~}}}}

\usepackage{amssymb}  

\usepackage[framemethod=tikz]{mdframed}  

\usepackage{dblfloatfix} 
\usepackage{placeins}

\newcommand*\samethanks[1][\value{footnote}]{\footnotemark[#1]}

\title{What Language Model Architecture and Pretraining Objective Work Best for Zero-Shot Generalization?}

\author{
\large \textbf{\underline{The BigScience Architecture \& Scaling Group}} \vspace{0.8cm}\\
\textbf{Thomas Wang$^1$\thanks{Equal contribution.} \hspace{0.5cm} Adam Roberts$^2$\samethanks} \vspace{0.3cm}\\
\textbf{Daniel Hesslow$^3$ \hspace{0.5cm} Teven Le Scao$^1$ \hspace{0.5cm} Hyung Won Chung$^2$} \vspace{0.3cm}\\
\textbf{Iz Beltagy$^4$ \hspace{0.5cm} Julien Launay$^{3, 5}$\thanks{Equal supervision. \newline Individual contributions outlined in \cref{sec:sup-contributions}} \hspace{0.5cm} Colin Raffel$^{1}\samethanks$}
\vspace{0.5cm}\\
 $^1$ Hugging Face \hspace{0.4cm} $^2$Google \hspace{0.4cm} $^3$LightOn \vspace{0.2cm}\\
 $^4$Allen Institute for AI \hspace{0.4cm} $^5$LPENS, École Normale Supérieure
\vspace{0.2cm}
}

\begin{document}

\maketitle

\begin{abstract}
  Large pretrained Transformer language models have been shown to exhibit \textit{zero-shot generalization}, i.e.\ they can perform a wide variety of tasks that they were not explicitly trained on.
However, the architectures and pretraining objectives used across state-of-the-art models differ significantly, and there has been limited systematic comparison of these factors.
In this work, we present a large-scale evaluation of modeling choices and their impact on zero-shot generalization.
In particular, we focus on text-to-text models and experiment with three model architectures (causal/non-causal decoder-only and encoder-decoder), trained with two different pretraining objectives (autoregressive and masked language modeling), and evaluated with and without multitask prompted finetuning. 
We train models with over 5 billion parameters for more than 170 billion tokens, thereby increasing the likelihood that our conclusions will transfer to even larger scales.
Our experiments show that causal decoder-only models trained on an autoregressive language modeling objective exhibit the strongest zero-shot generalization after purely unsupervised pretraining.
However, models with non-causal visibility on their input trained with a masked language modeling objective followed by multitask finetuning perform the best among our experiments.
We therefore consider the adaptation of pretrained models across architectures and objectives.
We find that pretrained non-causal decoder models can be adapted into performant generative causal decoder models, using autoregressive language modeling as a downstream task.
Furthermore, we find that pretrained causal decoder models can be efficiently adapted into non-causal decoder models, ultimately achieving competitive performance after multitask finetuning. Code and checkpoints are available at \href{https://github.com/bigscience-workshop/architecture-objective}{https://github.com/bigscience-workshop/architecture-objective}.
\end{abstract}

\newpage

\section{Introduction}
\label{sec:intro}
\begin{figure*}[t]
    \centering
    \includegraphics[width=\textwidth]{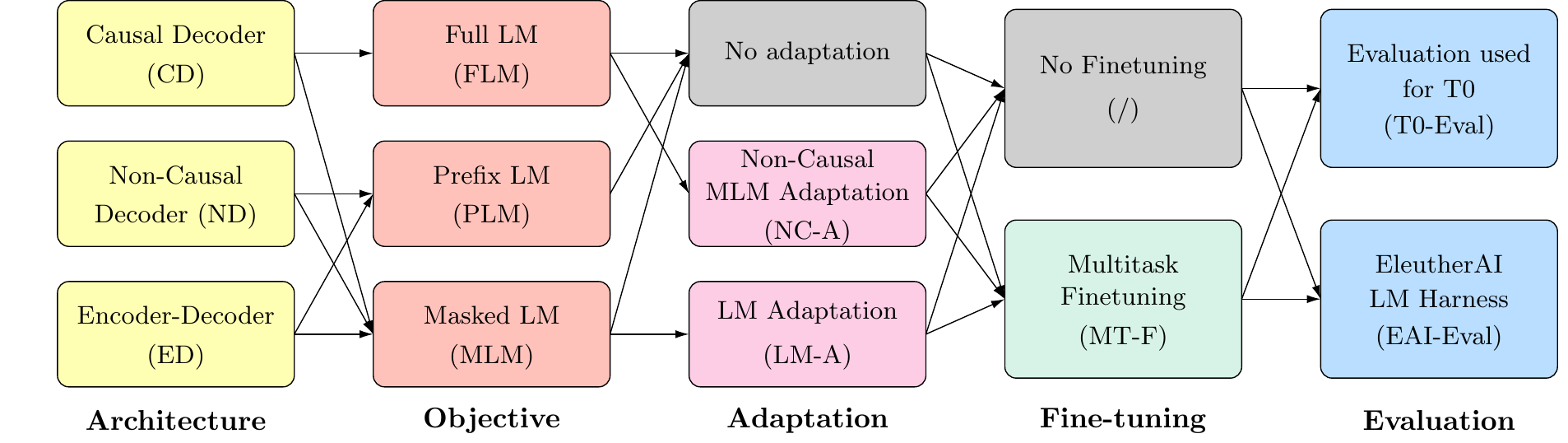}
    \caption{\textbf{We perform an extensive study of architecture, objective, adaptation, and finetuning impact on zero-shot generalization.} With 6 pre-trained models of over 5 billion parameters each trained on 168 billion tokens, adaptations on up to 100 billion tokens, multi-task finetuning on 13 billion tokens, and evaluation on 30 tasks from 2 different benchmarks with varied prompts, our study is the largest to date on the influence of modeling decisions on zero-shot generalization.}
    \label{fig:block}
    \vspace{-0.3cm}
\end{figure*}

Large language models (LLMs) pretrained on unstructured text data have been shown to be capable of performing a wide variety of text processing tasks without additional training.
This ability has been referred to as \evaluation{zero-shot generalization} since these models are typically pretrained with a self-supervised objective that is not specific to a downstream task.
Zero-shot generalization is particularly useful because it does not require any additional data or training in order to enable the model to perform a given task.
As such, there has been an explosion of work on developing LLMs and training techniques that produce strong zero-shot generalization \citep{gpt3, gpt-j, du2021glam, lin2021few, gopher, chinchilla, palm}.
One recent line of work \citep{T0, FLAN, ZeroPrompt} has demonstrated that adding an explicit \finetuning{multitask finetuning stage} on an ensemble of prompted tasks after pretraining can significantly boost the zero-shot capabilities of LLMs.

Modern LLMs are based on the \architecture{Transformer architecture} \citep{vaswani2017attention}.
While the original Transformer included a separate encoder that processes input text and a decoder that generates target text, most recent LLMs are causal decoder-only (CD) models trained to autoregressively predict a text sequence \citep{prefixliu, gpt, al2019character}.
In contrast with this trend,~\citet{T5} has shown that encoder-decoder (ED) models outperform decoder-only LLMs for transfer learning (i.e.\ where a pretrained model is finetuned on a single downstream task). Non-causal decoders (ND) \citep{prefixliu, prefixdong} use a modified attention mask to bridge the gap between decoder-only and encoder-decoder models. However, they have seen limited adoption.
Recently, \citet{T0} proposed a multitask finetuned encoder-decoder LLM that outperforms decoder-only models on zero-shot generalization, despite being an order of magnitude smaller. Concurrent work also demonstrated this approach with a decoder-only model \citep{FLAN}. This begs the question as to whether an encoder-decoder or a decoder-only would be a better choice for zero-shot generalization, especially if used in conjunction with multitask finetuning.

Transformer models can be trained with a variety of \objective{unsupervised training objectives}. Typically, decoder-only LLMs are pretrained using a \emph{full} language modeling (FLM) objective with a loss computed on all tokens \citep{dai2015semi, gpt}, and encoder-decoder models with a \emph{masked} language modeling (MLM) objective \citep{taylor1953cloze, bert}, such as span corruption \citep{raffel2019t5, joshi2020spanbert}.
It has repeatedly been shown that an MLM objective produces a better pretrained model for subsequent supervised finetuning in transfer learning settings~\citep{bert,lample2019cross,voita2019bottom,T5}.
The frequent use of the standard full language modeling objective nonetheless could be attributed to the fact that it lends itself to straightforward application of the model to many downstream tasks \citep{gpt2}. 
Still, the effectiveness of MLM in the transfer learning setting suggests it could create LLMs that are better suited to multitask finetuning.
Notably, the T0 model of \citet{T0} used an MLM pretraining objective, which may have contributed to its strong performance relative to larger models trained with only an FLM objective. 
Recently, \citet{lester2021power} also proposed introducing an \adaptation{adaptation stage} (i.e. extending pretraining but with a different objective) to enable an MLM model to perform prompted text generation tasks, bridging the gap across objectives.

These results indicate a need for a more systematic analysis of which architecture and pretraining objective pair produces LLMs with the strongest zero-shot generalization capabilities.
Past studies on architectures and objectives for language models \citep[e.g.][]{narang2021transfer, T5} have focused mainly on the transfer learning setting, with models that were orders of magnitude smaller than the current state-of-the-art.
Furthermore, recent results demonstrating the effectiveness of multitask finetuning raise the question of which architecture and pretraining objective is best suited to that promising setting.
Finally, novel adaptation practices also question the rigidity of these architecture and objective choices, and whether it is possible to efficiently convert pretrained models from one architecture to another. We propose to fill this gap with the following contributions.

\textbf{Large-scale systematic study.} We undertake a study of \architecture{architecture} and \objective{pretraining objective} combinations for LLMs with a focus on \evaluation{zero-shot generalization}. We consider decoder-only and encoder-decoder models using standard, prefix, and masked language modeling, spanning six $\langle$architecture, objective$\rangle$ pairs. We also evaluate performance with and without \finetuning{multitask finetuning}. In hopes of producing insights that transfer to very large models, we undertake our experiments at large scale: we train models with 5 billion parameters (11 for encoder-decoder) on 168 billion tokens, and perform multitask finetuning on 13 billion tokens. We base our zero-shot evaluation on the set of tasks used by \citet{T0} (T0-Eval) and the EleutherAI language model evaluation harness (EAI-Eval, \citet{eval-harness}), totalling 30 different datasets with varied prompt styles. \cref{fig:block} provides an overview of our study, \cref{sec:background} introduces background on the LLM architectures, objectives, training strategies, and evaluations considered, and \cref{sec:methods} details our methods.

\textbf{Multitask finetuning impacts architecture and objective choice.} We find that the popular recipe of a decoder-only model trained with a standard FLM objective performs best when zero-shot capabilities are measured immediately after pretraining, without any finetuning or adaptation.
However, after multitask finetuning, the results are the opposite: models pretrained with MLM perform significantly better and decoder-only models perform worse. These experimental results are discussed in \cref{sec:experiments}.

\textbf{Bridging across architectures and objectives with adaptation.}
This discrepancy motivates us to explore the practice of adaptation (i.e. extending the pretraining of a model with a different architecture/objective) as a way to efficiently obtain both a model suited to generative use cases and to mutitask finetuning.
We first consider \adaptation{language modeling adaptation}: adapting an MLM-trained non-causal decoder model by converting it to a causal decoder and extending its pretraining with an FLM objective. We find that using a pretrained model in this way speeds up convergence on the language modeling task by a factor 1.6x.
We then explore \adaptation{non-causal MLM adaptation}, starting from a causal decoder trained with an FLM objective, converting it to a non-causal decoder, and expanding its pretraining with an MLM objective. This form of adaptation produces a new version of the model suited for multitask finetuning, achieving second-best performance across our benchmarks. Convergence on the MLM task is sped-up by 3.3x, making this the most efficient approach to obtain two distinct models for generative tasks and multitask finetuning. We detail these results in \cref{sec:lm-adaptation}.


Accordingly, our results both confirm the validity of current standard practices, and help better understand the interplay between architecture, objective, multi-task fine-tuning, and zero-shot generalization. They also identify novel paths forward for efficiently obtaining better LLMs suited to either purely generative prompted usecases, or for multitask finetuning, as discussed in \cref{sec:conclusion}. 

\begin{figure*}[b]
    \centering
    \includegraphics[width = \textwidth]{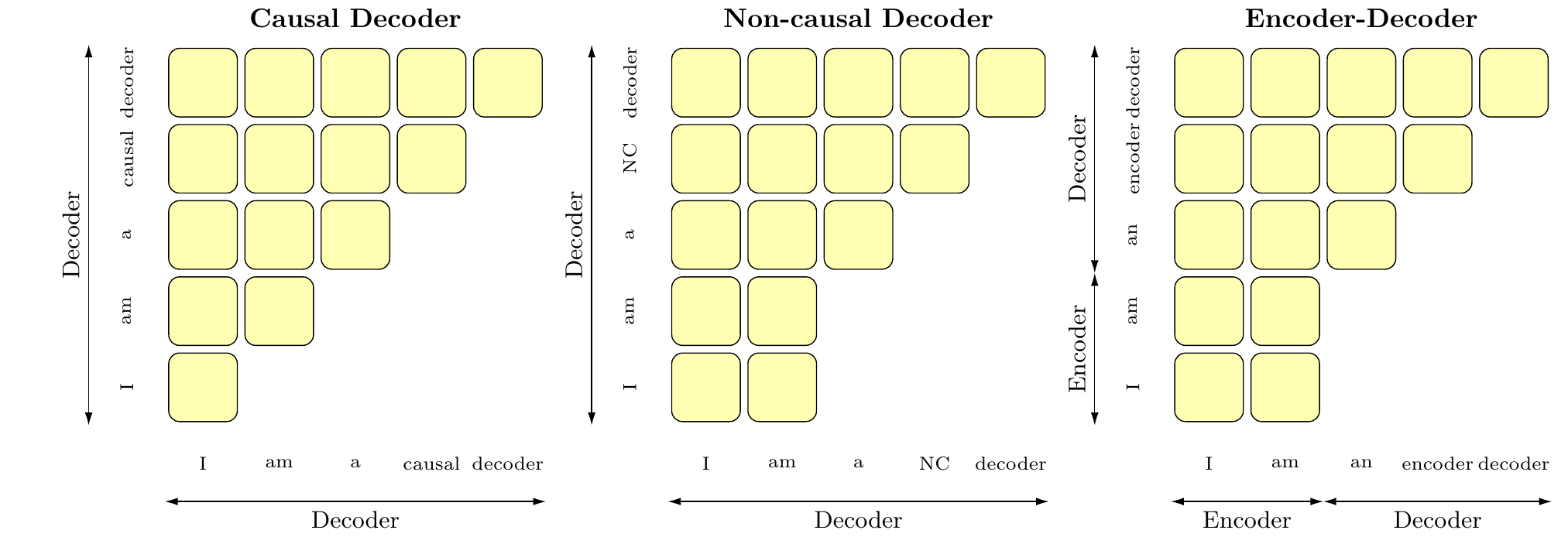}
    \caption{ \textbf{Attention patterns in a causal decoder, non-causal decoder, and encoder-decoder architecture}. In a causal decoder, each token attends to the previous tokens only. In both non-causal decoder and encoder-decoder, attention is allowed to be bidirectional on any conditioning information. For the encoder-decoder, that conditioning is fed into the encoder part of the model.}
    \label{fig:archs}
\end{figure*}
\section{Background}
\label{sec:background}
\subsection{Architectures}

\textbf{Transformer.} Virtually all state-of-the-art LLMs are based on the Transformer architecture \citep{vaswani2017attention}.
Due to its ubiquity, we only highlight a few relevant high-level characteristics. The main architectural unit of the Transformer is a Transformer \emph{block}, which consists of (at minimum) multi-headed self attention \citep{cheng2016long}, layer normalization \citep{ba2016layer}, a dense two-layer feedforward network, and residual connections \citep{he2016deep}.
A Transformer \emph{stack} is a sequence of such blocks.
In NLP applications, the Transformer ingests and outputs \emph{tokens}.
Since being introduced by \citet{vaswani2017attention}, various \architecture{architectural variants} of the Transformer have been proposed. A major difference between these architectures is the masking pattern applied to the provided \emph{inputs}, which act as contextual information for the model to make a prediction. \cref{fig:archs} showcases the attention masking patterns in the three architectural variants we consider.

\textbf{Encoder-decoder.}
As originally proposed, the Transformer consisted of two stacks: an encoder and a decoder.
The encoder is fed the sequence of input tokens and outputs a sequence of vectors of the same length as the input.
Then, the decoder autoregressively predicts the target sequence, token by token, conditioned on the output of the encoder.
To achieve this conditioning, the decoder includes cross-attention layers in each of its blocks, allowing the decoder to also attend to the output of the encoder. 
The self-attention layers in the decoder utilize a \emph{causal} masking pattern that prevents the model from attending to future tokens when predicting the output sequence (see \cref{fig:archs}, on the right).
We hereafter refer to this architecture as the \architecture{encoder-decoder~(ED)}.
Notable pretrained language models using an encoder-decoder architecture include BART \citep{bart} and T5~\citep{T5}.
T5 in particular was recently used as the foundation for the T0 model \citep{T0}, which leveraged large-scale multitask finetuning to achieve strong zero-shot generalization, outperforming decoder-only models an order of magnitude larger.

\textbf{Causal decoder-only.} Although the encoder-decoder is the original Transformer variant, most recent LLMs use a decoder-only architecture.
These models can be trained as a traditional language model~(i.e.\ to predict the next token in a sequence).
Decoder-only models have no independent means of processing or representing the input sequence and target sequence differently--all tokens are processed in an equivalent fashion, and, because of the causal masking pattern, conditioning is simply based on past tokens (see \cref{fig:archs}, on the left).
On the one hand, this means that the representation for any conditioning text is inherently weaker; on the other hand, it yields a simpler architecture that is naturally suited to a standard autoregressive next-step-prediction pretraining objective.
We refer to this architecture as \architecture{causal decoder-only (CD)}.
Most notably, the CD architecture makes up the backbone of the GPT series of models \citep{gpt, gpt2, gpt3} as well as many other recent record-breaking LLMs \citep{zeng2021pangu, kim2021changes, smith2022using, lamda, gopher, chinchilla, palm}.

\textbf{Non-causal decoder-only.}
To allow decoder-only models to build richer non-causal representations of the input/conditioning text, it has been proposed to simply modify the attention mask used.
Specifically, the self-attention masking pattern can be changed so that the region of the input sequence corresponding to conditioning information has a non-causal mask (i.e.\ attention in this region is not restricted to past tokens, see middle of \cref{fig:archs}), as in the encoder of an encoder-decoder architecture.  We refer to this architecture as \architecture{non-causal decoder-only (ND)}. 
Sometimes called a \emph{prefix} language model, this approach was introduced by \citep{prefixliu} and was later explored as an architectural variant by \citep{T5,wu2021yuan}. Despite single-task finetuning performance nearly on par with encoder-decoder models \citep{T5}, it has seen limited adoption in the literature.

\begin{figure*}[t!]
    \centering
    \includegraphics[width=0.9\textwidth]{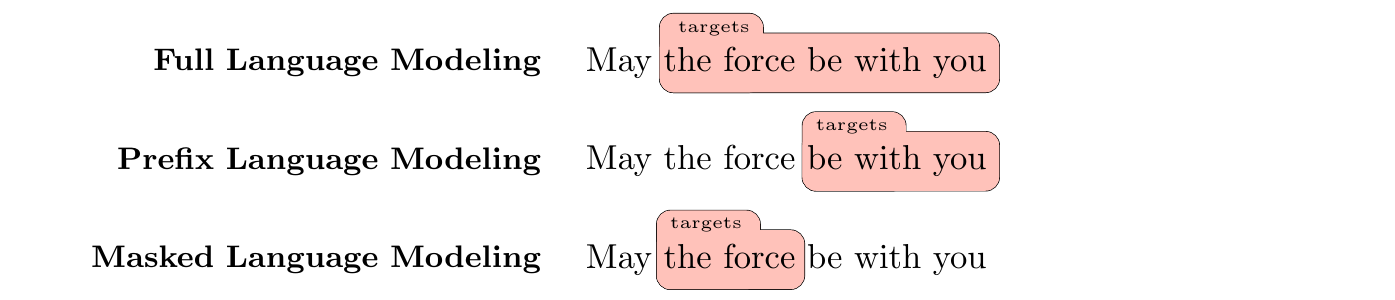}
    \caption{\textbf{Input and targets tokens in full, prefix, and masked language modeling training objectives.} For full language modeling, all tokens in a sequence are used during training. For prefix language modeling, we randomly select a prefix size, and hence only half of the tokens are used on average to derive the loss. At inference time, the prefix would be over the input/conditioning information. Finally, for masked language modeling, we mask 15\% of the tokens, in spans of 3 tokens on average. We use sentinel tokens to replace spans (not represented here), and the model outputs subsequently each sentinel followed by its prediction of the content masked by the sentinel.}
    \label{fig:objectives}
    \vspace{-0.4cm}
\end{figure*}

\textbf{Encoder-only.}
As an aside, we note that another popular architectural variant is to only use a Transformer encoder layer stack.
This model architecture underlies the ubiquitous BERT \citep{bert} and its derivatives.
However, this architecture is limited to producing the same number of tokens as it was fed as input, considerably limiting its applicability and making it only rarely used in the zero-shot setting \citep{ed_zeroshot}. We therefore omit it from consideration.

\textbf{Comparisons across architectures.} Decoder-only models process a single sequence consisting of the concatenation of the input and target text. On the other hand, in an encoder-decoder, the encoder processes only the input and the decoder processes only the target.
The total \textit{amount of computation} performed by an encoder-decoder will therefore be approximately equivalent to a decoder-only model when the encoder and decoder each have as many parameters as the entire decoder-only model (ignoring the cross-attention layers in the encoder-decoder). However, such an encoder-decoder will have twice the parameters of the decoder-only model, and hence twice the memory footprint. 

\subsection{Pretraining objectives}

An important step in building LLMs is pretraining, where the model is trained on a large, unlabeled dataset via self-supervision. The choice of \objective{pretraining objective} can have significant impact on the downstream usability of the LLM, and we therefore include objective choice as a factor in our empirical study. Figure \ref{fig:objectives} outlines the input and target tokens for the pretraining objectives considered.


\textbf{Language modeling.}
Since the advent of GPT-2 \citep{gpt2}, large decoder-only models have generally been pretrained with an autoregressive language modeling objective \citep{gpt3, wu2021yuan, gopher}. Given previous tokens, the model is tasked with predicting the following one. We refer to this as \objective{full language modeling (FLM)}.
This objective is particularly efficient during pretraining: all tokens in a sequence can generate a loss signal in parallel. At inference time, the model is iteratively asked to predict the next token. 

\textbf{Prefix language modeling.}
For encoder-decoder and non-causal decoder-only models to perform language modeling, one can define a prefix where the attention mask is allowed to be non-causal. Similar to standard language modeling, the model is tasked to predict each token outside the prefix given all previous tokens. We hereafter refer to this objective as \objective{prefix language modeling (PLM)}. Loss on the prefix is ignored as tokens in the prefix can attend to their targets. For inference, the prefix is naturally the input text; during pretraining, it is usually chosen at random for each sample.

\textbf{Masked language modeling.}
Encoder-only models, such as BERT \citep{bert}, have typically been pretrained with a masked language modeling objective. Tokens or spans of tokens in the input text are replaced with a special mask token and the model is trained to predict the missing tokens. \citet{T5} introduced a version of this objective adapted to text-to-text models in the form of \emph{span corruption}: sentinel tokens are used to flag masked spans of short random lengths, and, after processing the masked input, the model outputs the sentinels followed by their respective predicted content. We refer to this approach as \objective{masked language modeling (MLM)}.

\subsection{Model adaptation}

Adaptation extends pretraining with a different objective and/or architecture. In contrast with finetuning, no new \emph{downstream} data is used, only additional pretraining data. \adaptation{Language modeling adaptation (LM-A)} takes a model pretrained with MLM and extend its training with PLM or FLM. It has been used to convert encoder-decoder models pretrained with MLM, such as T5, into better generative models. Notably, it is used as a first step before prompt tuning \citep{lester2021power} and also to prepare the model before multitask finetuning in T0 \citep{T0}. When we perform language modeling adaptation on a non-causal decoder-only model, we convert it into a causal decoder-only by simply switching the attention mask.
Furthermore, we propose to study the opposite adaptation: starting from a causal decoder pretrained with FLM, we cast the model into a non-causal decoder (again by switching the attention mask) and we extend pretraining with MLM. We call this approach \adaptation{non-causal MLM adaptation (NC-A)}; to our knowledge, this is an entirely novel practice.

\subsection{Multitask finetuning}

Modern pretraining corpora are typically massive preprocessed generalist webcrawls \citep{oscar, T5}, collected with no explicit regard for downstream tasks--although adding curated high-quality cross-domain data has been proposed as a path towards better zero-shot generalization \citep{pile, anonymous2022what}. Recently, \citet{T0} (on an encoder-decoder model trained with MLM) and \citet{FLAN} (on a causal decoder-only model trained with FLM) explored the potential of explicitly finetuning the model to solve multiple tasks in order to bolster zero-shot generalization. This is done by finetuning the model on a dataset of prompted tasks (i.e.\ in a natural language format, leveraging prompt templates applied over many datasets), which ultimately improves zero-shot performance over purely unsupervised pretraining. We refer to this as \finetuning{multitask finetuning (MT-F)}, and use the openly available datasets and prompts developed for T0.

\subsection{Zero-shot evaluation}

\citet{gpt2} first demonstrated that LLMs display zero-shot capabilities: given sufficient scale, language models are able to perform many tasks without having explicitly accessed any supervised samples. Zero-shot use of language models relies on a technique called \textit{prompting}, where tasks are formulated in a natural language format (in accordance with the pretraining objective). The template applied to each example to convert it to this format is called the prompt. Unfortunately, models can exhibit significant sensitivity to the wording of the prompt \citep{T0}, and it can be difficult to diagnose whether poor performance is a prompt- or model-related problem. 

Zero-shot capabilities are of increasing interest in the community, as evidenced by most record-breaking LLMs only reporting zero/few-shot results \citep{gpt3, smith2022using, gopher, palm}. There are many reasons why zero-shot use is gaining such traction: it does not require any labeled examples, it removes the complexity of model finetuning and deployment, and it also tests generalization to unseen tasks.

We rely on two evaluation benchmarks aggregating prompts across NLP tasks, totalling 30 tasks: the \evaluation{EleutherAI LM Evaluation Harness (EAI-Eval)} \citep{eval-harness}, which reimplements the prompts from~\citet{gpt3} and is aimed at evaluation of FLM-trained causal decoder-only models, and the \evaluation{evaluation set from T0 (T0-Eval)} \citep{T0}. Note that EAI-Eval only includes one prompt per task, whereas performance on T0-Eval is averaged over many prompts. Hence, when reporting performance on T0-Eval, we report a spread across prompts, giving an indication of the impact of the choice of prompt on performance.

\section{Methods}
\label{sec:methods}
To better understand how architecture, pretraining objective, multitask finetuning, and possible adaptations influence zero-shot performance, we undertake a systematic large-scale study. We pretrain all possible $\langle$architecture, objective$\rangle$ pairs on 168 billion tokens from C4, consider intermediate multitask finetuning, and finally evaluate zero-shot performance. We also study the possibility of using adaptation to efficiently transfer the benefits from one architecture/objective to another.

\paragraph{Compute budget guideline.} Different architectures and objectives come with different compute trade-offs. We aim to make the training budget similar across all models, using $\sim15$ petaflops-days for pretraining (for a total of 830,000 TPUv4-hours over the study, see \cref{sec:sup-carbon} for details). We do not take into account memory use: typical use cases are compute-bound by the available GPU/TPU-hours. We note that the encoder-decoder ends up with twice the memory footprint. 

\paragraph{Resources and implementation.} We run all computation on Google Cloud TPUv4s, using T5X \citep{roberts2022t5x}, leveraging JAX \citep{JAX} and Flax~\citep{flax}. 

\subsection{Architecture}

\begin{center}
\end{center}
\begin{table}[b]
\caption{\textbf{Shared architecture for all models trained.} Encoder-decoder architectures are doubled in size to obtain a pretraining compute budget similar to the decoder-only architecture.}
\label{tab:architecture}
\begin{center}
\begin{small}
\begin{tabular}{lcc} 
 \toprule
 & \multicolumn{2}{c}{\sc{Models architecture}} \\
 & Decoder-only & Encoder-decoder \\
 \midrule
 Parameters & 4.8B & 11.0B \\
 Vocabulary & \multicolumn{2}{c}{32,128} \\
 Positional embed. & \multicolumn{2}{c}{T5 relative} \\
 Embedding dim. & \multicolumn{2}{c}{4,096} \\ 
 Attention heads & \multicolumn{2}{c}{64}  \\
 Feedforward dim. & \multicolumn{2}{c}{10,240} \\
 Activation & \multicolumn{2}{c}{GEGLU \citep{shazeer2020glu}}  \\
 Layers & 24 & 48 \\
 Tied embeddings & \multicolumn{2}{c}{True} \\
 Precision & \multicolumn{2}{c}{\texttt{bfloat16}}  \\
 \bottomrule
\end{tabular}
\end{small}
\end{center}
\end{table}

\vskip -0.3in
We consider \architecture{causal decoder (CD)}, \architecture{encoder-decoder (ED)}, and \architecture{non-causal decoder (ND)} architectures. All models share the basic configuration outlined in Table \ref{tab:architecture}.
For fair comparison across architectures, we aim to approximately match pretraining compute budget; accordingly, our encoder-decoder models have twice as many layers as the decoder-only models. This results in encoder-decoder models with 11B parameters and decoder-only models with 4.8B parameters. 
We note that due to the cross-attention layers, encoder-decoder models are approximately $\sim 10$\% more computationally expensive to run than the decoder-only models we consider.

\subsection{Pretraining}

We consider \objective{full language modeling (FLM)}, \objective{prefix language modeling (PLM)}, and \objective{masked language modeling (MLM)} (specifically, the span corruption objective of \citet{T5}). The choice of language modeling objective depends on the architecture: the causal decoder uses either FLM or MLM, while the non-causal decoder and the encoder-decoder use either PLM or MLM. 

All of our models are pretrained on 168 billion tokens of the C4 dataset from \citet{T5}.
We use the Adafactor \citep{shazeer2018adafactor} optimizer with an inverse square root learning rate schedule, training on batches of 2,048 sequences of length 626 tokens (for a total of 131,072 training steps). Detailed pretraining hyperparameters can be found in Table \ref{tab:pretraining}: we based elements of our pretraining setup (such as Adafactor, GEGLU, and the use of an auxiliary $Z$ loss $\mathcal{L}(Z) = 10^{-4} * \log^2(Z)$ to stabilize training~\citep{palm}) on the popular T5.1.1 recipe.

To operate with a fixed compute budget, we match the amount of tokens seen during pretraining (which corresponds to the total computational cost), not the number of tokens trained on (i.e.\ on which a loss is calculated). Full language modeling computes a loss on all the tokens it sees, whereas prefix language modeling cannot train on the tokens in its prefix: on average, it will train on half as many tokens as full language modeling. We consider these to be inherent trade-offs in efficiency between training objectives.
We concatenated and sampled text from documents in such a way that there was virtually no padding during pretraining. More specifically to each objective:

\begin{itemize}
    \item For full language modeling, the loss is computed for all 626 token in each sequence in parallel, making for the most efficient configuration (100\% of tokens are trained on).
    \item For prefix language modeling, we select a random split point in [1, 626], which we use as the prefix length of one example and the suffix length for another, packing them together to avoid padding (using appropriately masked attention), and computing the loss only on the suffixes (50\% of tokens on average). See \cref{sec:sup-tpu} for implementation details on TPUs.
    \item For masked language modeling, 15\% of input tokens are masked with an average span length of 3 (as used by \citet{T5}), such that there are approximately 512 input and 114 target tokens, with the loss computed only on the targets (18\% of tokens on average).
\end{itemize}

\begin{table}[b]
\caption{\textbf{Pretraining and multitask finetuning configurations for all models trained.} Pretraining lasts for 168 billion tokens, while multitask finetuning is done for 13 billion tokens.}
\label{tab:pretraining}
\begin{center}
\begin{small}
\begin{tabular}{lccc}
 \toprule
 & \multicolumn{1}{c}{\sc{Pretraining}} &  \multicolumn{1}{c}{\sc{Multitask finetuning}}\\
 \midrule
 Dataset & C4 & T0-Train\\ 
 Steps & 131,072 & 10,000 \\ 
 Batch size in tokens & 1,282,048 & 1,310,720 \\
 Optimizer & \multicolumn{2}{c}{Adafactor(decay\_rate=0.8)}  \\
 LR schedule & $\frac{1}{\sqrt{\max(n, 10^4)}}$ & fixed, 0.001\\
 Dropout & 0.0 & 0.1  \\
 z loss & \multicolumn{2}{c}{0.0001}  \\
 Precision & \multicolumn{2}{c}{\texttt{bfloat16}} \\
 \bottomrule
\end{tabular}
\end{small}
\end{center}
\end{table}

\subsection{Multitask finetuning}

Drawing from recent work demonstrating that multitask finetuning improves zero-shot performance, we also evaluate our models after \finetuning{multitask finetuning (MT-F)}, following the procedure used for the T0 model of \citet{T0}. Our goal is to better disambiguate the influence of architecture and objective in this relatively nascent practice. For example, we note that T0 and FLAN are significantly different in the architecture and objective used (encoder-decoder with MLM and causal decoder with FLM, respectively). We hope our experiments can help lend insight into which of these design choices is more effective for enabling zero-shot generalization after multitask finetuning.

After pretraining, we create multitask versions of our models by finetuning on the T0 training dataset mixture from \citet{T0} (not T0+ or T0++) for 13 billion tokens. Our finetuning configurations follow those used for T0 (see \cref{tab:pretraining} for details), and note that we found dropout to significantly impact zero-shot generalization (see \cref{sec:sup-dropout} for a comparison with and without dropout). 
We refer readers to \citet{T0} for further information about this multitask finetuning procedure.

One significant departure to note from the approach of \citet{T0} is that we do not perform language modeling adaptation first before multitask finetuning. Preliminary results (see \cref{sec:sup-res} in the Appendix) did not show any systematic improvement from performing language modeling adaptation, so we omitted this step. This is consistent with the finding from \citet{lester2021power} that language modeling adaptation is not necessary before prompt tuning for large models.

\subsection{Evaluation}

We use two zero-shot evaluation benchmarks to assess our models. First, we use the \evaluation{same set of tasks, datasets, and prompts as was used to evaluate T0 (T0-Eval)} \citep{T0}, and second, the \evaluation{EleutherAI LM evaluation harness (EAI-Eval)} \citep{eval-harness}. 
The EAI prompts attempt to replicate the evaluation set of \citet{gpt3}. The prompts of T0 were built to be ``human understandable'' and were originally used in conjunction with an encoder-decoder model. See \cref{sec:sup-eval} for a detailed list of tasks, and the overlap between T0-Eval, EAI-Eval, and T0-Train.

T0-Eval provides multiple prompts per task, whereas EAI-Eval provides only one prompt per task. Accordingly, for T0-Eval, we take the median accuracy over all prompts for each task and then average across all 11 datasets. For EAI-Eval we simply average the accuracy obtained on each of the 31 datasets. Note that because these are aggregated zero-shot benchmarks, variations of even a percent can hide significant differences on a single task. 

All but one task in T0-Eval (StoryCloze) are also in EAI-Eval. Some of the datasets in EAI-Eval are also in the T0 training datasets: GLUE-MRPC, GLUE-QQP, and SciQ. We also did not check for contamination from C4, but given the fact that all models would have the opportunity to memorize the tasks leaked in C4, we believe it does not impact our evaluation. For additional discussion of the overlap between T0 and C4, we refer readers to the original T0 paper~\citet{T0}.

We perform evaluation of model checkpoints at 42B, 84B, and 168B tokens. We note that the random baselines are $\sim33$\% for EAI-Eval and $\sim42$\% for T0-Eval. The complete set of results across all checkpoints obtained through this study is made available in \cref{sec:sup_complete-results}.

\section{Experiments}
\label{sec:experiments}
\subsection{After unsupervised pretraining only}
\label{sec:pretraining-only}

We are first interested in the architecture and objective achieving the best zero-shot performance after unsupervised pretraining only. For this, we only consider the full/prefix language modeling objectives since masked language modeling does not yield a model appropriate for zero-shot prompted evaluation on its own. This is validated with early checkpoints in \cref{sec:sup-res}.

We present our main full/prefix language modeling pretraining results in \cref{tab:lm-results}. On both our evaluation benchmarks, the causal decoder architecture systematically outperforms the other architectures when using language modeling pretraining alone. The non-causal decoder remains within a percent of the causal decoder performance, but the encoder-decoder performance lags far behind. Finally, we note that the performances on T0-Eval are close to the random baseline, while performance differences on EAI-Eval are significant enough to make comparison across experiments.

\begin{mdframed}
\textbf{Finding 1.} \architecture{Causal decoder-only} models pretrained with a \objective{full language modeling} objective achieve best zero-shot generalization when evaluated immediately after \neutral{unsupervised pretraining}, in line with current common practices for large language models.
\end{mdframed}

\begin{table}[b]
\caption{
\textbf{After full or prefix language modeling pretraining, the causal decoder (FLM) exhibits the best zero-shot generalization abilities, followed closely by the non-causal decoder (PLM).}
Average accuracy on EAI-Eval and T0-Eval after pretraining for 168B tokens. MLM pretraining is not considered here, as the models produced are not suitable for direct use in a zero-shot setting. Note that performance on T0-Eval remains close to random baseline. 
\neutral{Best for each benchmark.}
}
\label{tab:lm-results}
\begin{center}
\begin{small}
\begin{tabular}{@{}lcc@{}}
\toprule
                   & \sc{EAI-Eval} & \sc{T0-Eval} \\ \midrule
Causal decoder     & \cellcolor{neutral}\textbf{44.2}                            & \cellcolor{neutral}\textbf{42.4}                             \\
Non-causal decoder & 43.5                            & 41.8                             \\
Encoder-decoder    & 39.9                            & 41.7                             \\ \midrule
Random baseline & 32.9 & 41.7 \\ \bottomrule
\end{tabular}
\end{small}
\end{center}
\end{table}

\subsection{After multitask finetuning}

We now focus on the relatively new practice of multitask finetuning, where there has not yet been any systematic study of the influence of the architecture and training objective.
Notably, the two main papers advocating this practice use completely different approaches: \citet{T0} finetunes an encoder-decoder model pretrained with span corruption, whereas \citet{FLAN} finetunes a decoder-only pretrained with full language modeling.
It is not immediately clear which approach is more natural: while decoder-only models trained with full language modeling are better at zero-shot generalization (as evidenced in \cref{sec:pretraining-only}), encoder-decoder models and masked language modeling pretraining have been shown to perform significantly better after finetuning \citep{T5}.
We therefore evaluate every architecture and objective combination after multitask finetuning.

\begin{figure}[t]
    \centering
    \includegraphics[width=0.9\textwidth]{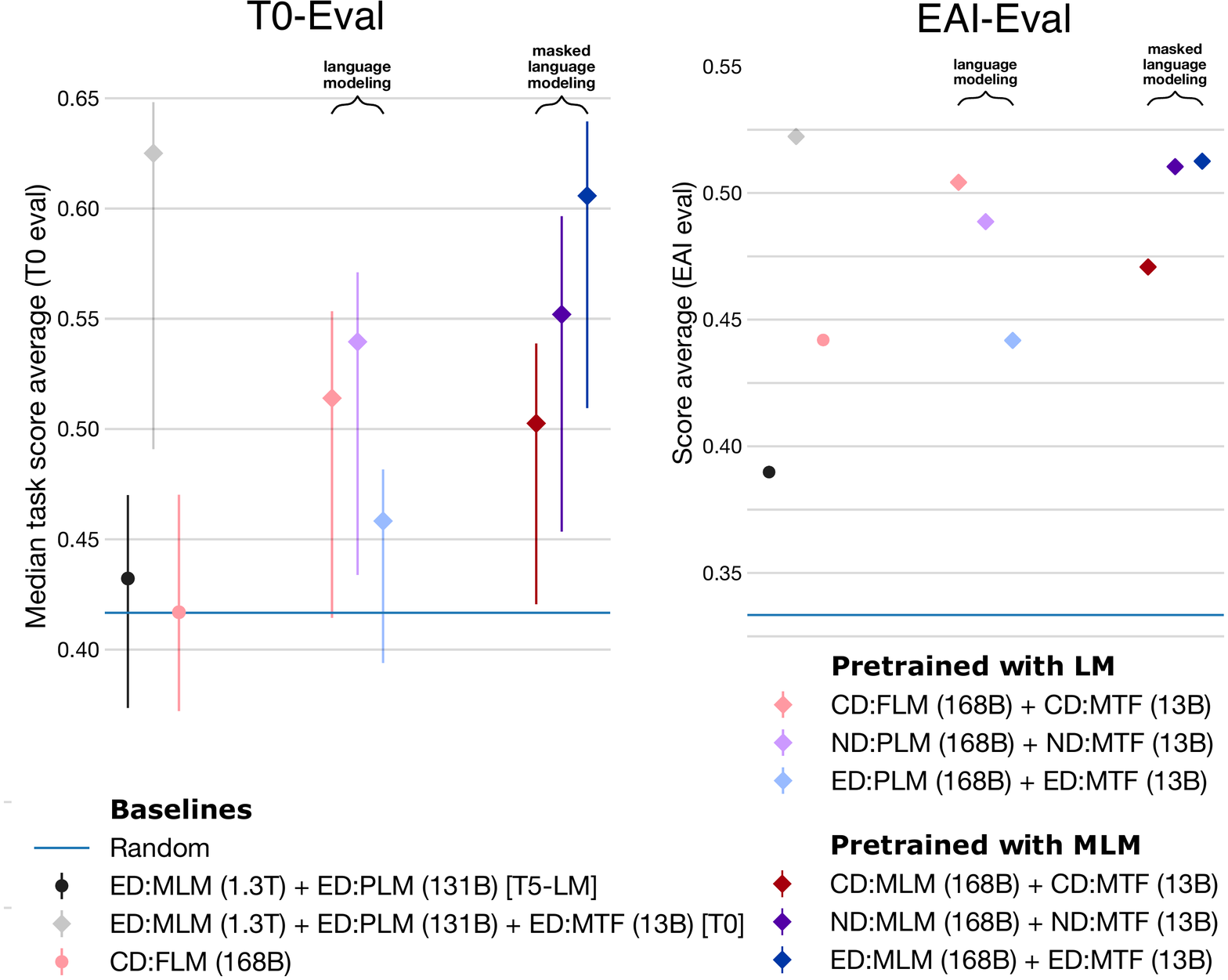}
    \caption{\textbf{When considering multitask finetuning, the encoder-decoder pretrained with MLM significantly outperforms other models, with the non-causal decoder pretrained with MLM a close second on EAI-Eval.} Detailed mean performance and spread compared to baselines on T0-Eval (left) and EAI-Eval (right) after multitask finetuning on T0 training set. The T5-LM and T0 results are taken from \citet{T0} and utilize the original T5.1.1-XXL model, which was pretrained for 7.6$\times$ the number of tokens used in this study. Other models are evaluated after pretraining with 168B  tokens and multitask finetuning on T0 training set for 13B tokens.}
    \label{fig:flagship_models}
\end{figure}

Our results are outlined in Figure \ref{fig:flagship_models}. The encoder-decoder pretrained with span corruption offers the best performance after multitask finetuning. Specifically, on EAI-Eval, the best performance is achieved by the encoder-decoder with MLM, and the non-causal decoder with MLM comes in a close second. However, the difference is more significant on T0-Eval, where the encoder-decoder with MLM pretraining outperforms other models by a large margin. Finally, encoder-decoder pretrained with PLM and causal decoder with MLM achieve significantly worse performance than other models. These results are consistent across all levels of pretraining (see early checkpoints in Appendix \ref{sec:sup-eval}).


\begin{mdframed}
\textbf{Finding 2.} \architecture{Encoder-decoder} models trained with \objective{masked language modeling} achieve the best zero-shot performance after \finetuning{multitask finetuning}. More broadly, approaches that perform well in the single-task finetuning setting perform well on multitask finetuning.
\end{mdframed}

\subsection{Influence of the tasks and prompts used for zero-shot evaluation}

Although the datasets considered in EAI-Eval and T0-Eval have significant overlap (10 out of 11 T0 tasks are in EAI-Eval), the prompts are always different between the two benchmarks.
The EAI prompts for these datasets were taken from \citet{gpt3}, who hand-tuned them to maximize performance of the GPT-3 models. In contrast, the T0-Eval prompts were sourced through a community effort with prompt diversity and naturalness as primary goals without any regard for model performance \citep{T0}. Consequently, on each task, the EAI prompt has higher performance than the average T0 prompt for all models and tends to be on par with the best T0 prompt. The difference is most pronounced for causal decoder language models without multitask finetuning, likely because this is the setting GPT-3 prompts were optimized for. This is reflected in the structure of the prompts, which tend not to explain the task to the reader like T0 evaluation prompts do. Instead, they attempt to reformulate it as close to language modeling as possible.

\begin{figure}[t]
    \centering
    \includegraphics[width=\textwidth]{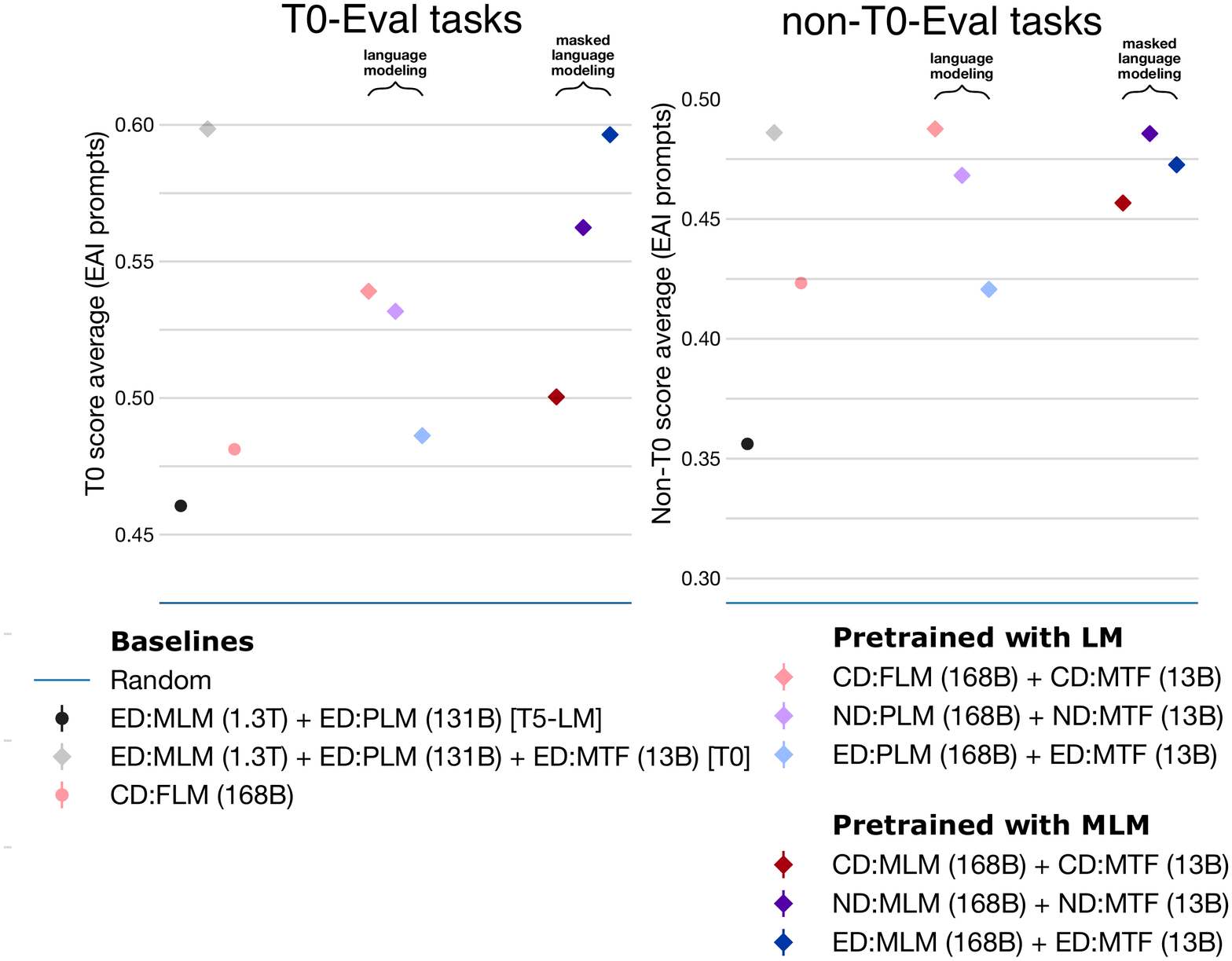}
    \caption{\textbf{Decoder-only models perform better on EAI-Eval because of specific tasks, not because of differences in prompts used.}
    Zero-shot performance on T0-Eval tasks (left) and non-T0-Eval tasks (right), both using EAI-Eval prompts to control for the influence of T0-Eval prompts.}
    \label{fig:task_comparison}
\end{figure}

In addition to this base performance discrepancy, EAI-Eval has less discrepancy between encoder-decoder models and the rest, and better performance for autoregressive decoder models.
We untangle the effect of the difference in prompts and the different task sets by separately comparing performance on tasks thare in T0-Eval and those that are not, while always using EAI-Eval prompts, as shown in \cref{fig:task_comparison}. 
The set of EAI-Eval tasks considered in T0-Eval seems to lend itself better to encoder-decoder models than the rest. On non-T0-Eval tasks, in contrast, causal decoder performance shoots up dramatically, although a lot of the difference is driven by LAMBADA \citep{lambada}, a language modeling task. Nevertheless, we note that when considering wide and varied task aggregates, our high-level findings are mostly consistent across evaluation settings--although specific tasks, such as LAMBADA, may indeed favor a specific architecture and objective combination.

\section{Can models be adapted from one architecture/objective to another?}
\label{sec:lm-adaptation}
Our experimental study has led us to conclude the optimal architecture and objective choice for zero-shot performance depends on whether or not the model will ultimately undergo multitask finetuning: while a decoder-only model trained with full language modeling achieves the best zero-shot performance after unsupervised pretraining only, an encoder-decoder with masked language modeling is best once multitask finetuning is applied. This is inconvenient, as the multitask finetuned encoder-decoder model may not be suitable for many open-ended generative tasks that the decoder-only model excels at, while the decoder-only model will not be the best at many zero-shot tasks.

In this section, we attempt a compromise between the two options above. We study the practice of adaptation: extending pretraining with a different architecture and/or objective. Our end-goal is to efficiently obtain two distinct models: one that leverages multitask finetuning to maximize zero-shot performance, and another that can be used as a high-quality language model.

\paragraph{Language modeling adaptation (LM-A).} First, we propose to pretrain a non-causal decoder model with an MLM objective and then further train the model as a causal decoder with a FLM objective (\emph{language modeling adaptation}). This conversion is simple, as the parameters and overall architecture can be kept the same, and only the attention mask needs to be switched. We note that we also attempted this adaptation from the decoder portion of an encoder-decoder model, but it performed significantly worse than training from scratch, as discussed in \cref{sec:sup-ec-adapt}. 

\begin{figure}[t]
    \centering
    \includegraphics[width=0.49\textwidth]{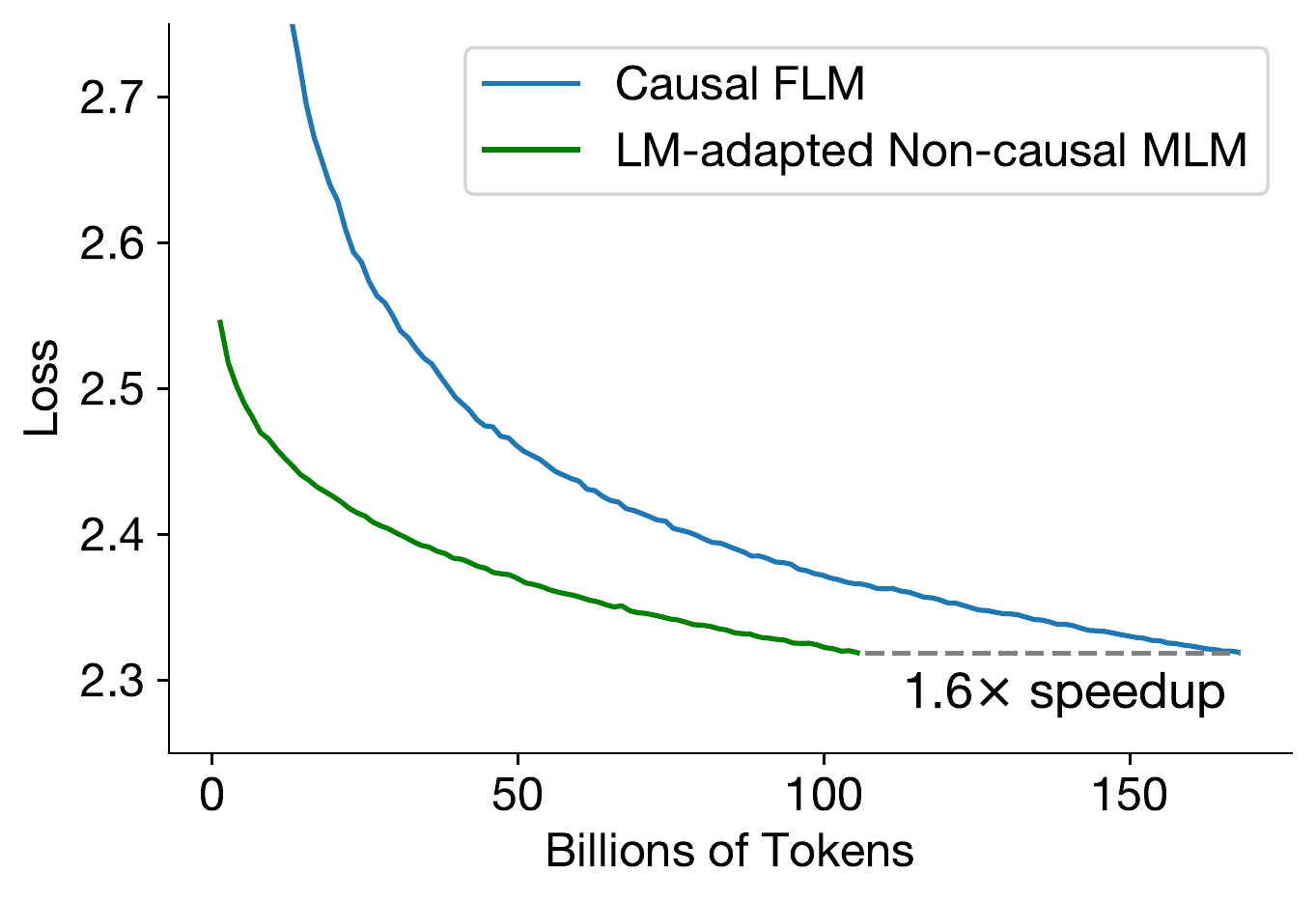}
    \includegraphics[width=0.49\textwidth]{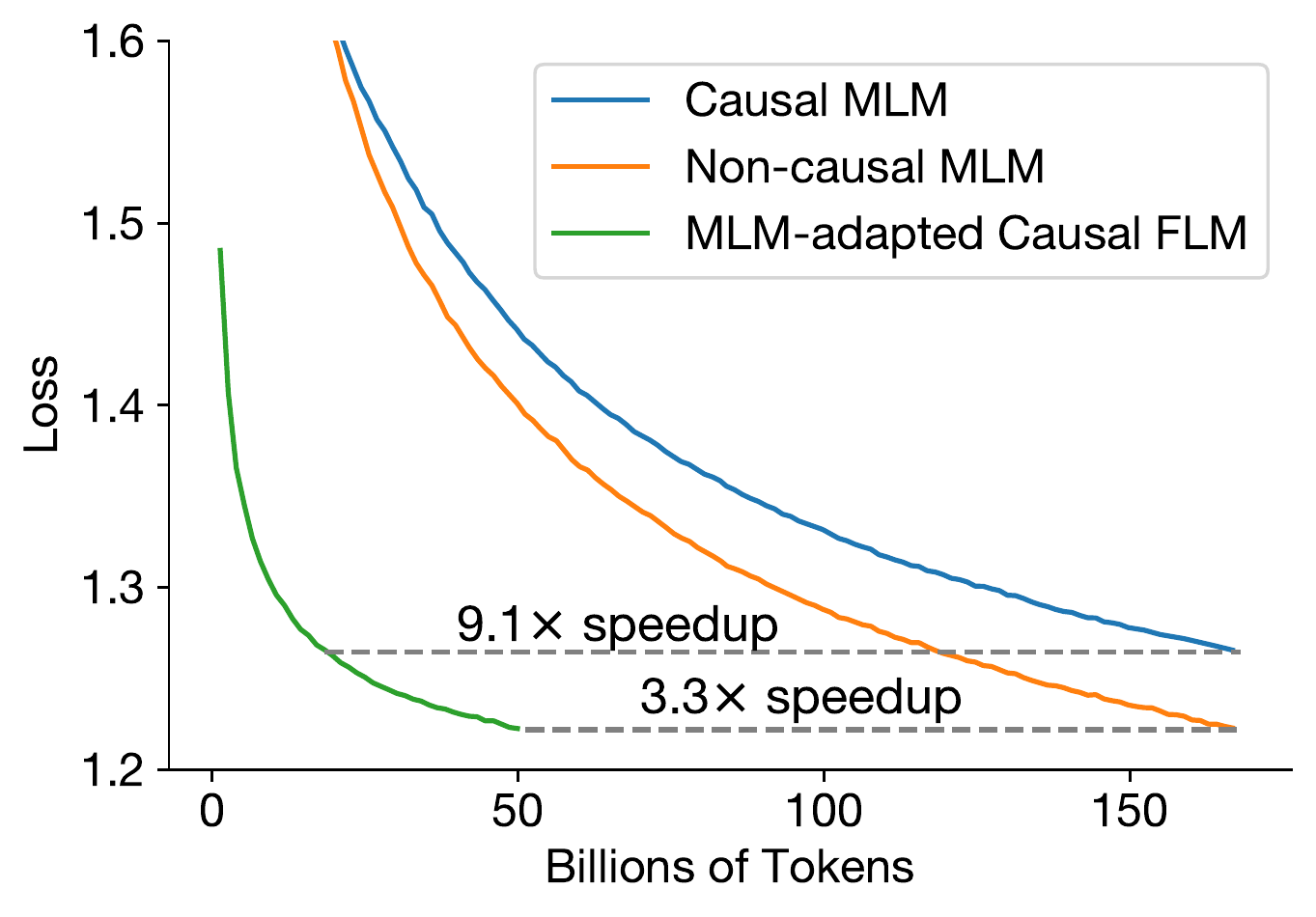}
    \caption{\textbf{Adaptation can efficiently convert non-causal decoder-only models pretrained with MLM into causal decoder-only models with FLM (left), and vice-versa (right).} Validation loss when adapting decoder-only models to different architecture/objectives following pretraining. Left: A causal decoder-only pretrained with FLM from scratch (Causal FLM) compared to a model being adapted with FLM \emph{following} 168B pretraining tokens as a non-causal masked language model (LM-adapted Non-causal MLM). The adaptation requires $63\%$ of the tokens (1.6$\times$ speedup) versus training from scratch. Right: Causal and non-causal decoder-only masked language models (Causal MLM, Non-causal MLM) trained from scratch compared to a model being adapted to a non-causal MLM \emph{following} 168B pretraining tokens as a causal FLM (MLM-adapted Causal FLM). The adaptation requires $30\%$ of the tokens (3.3$\times$ speedup) versus training the non-causal MLM from scratch.}
    \label{fig:adapt-speedup}
\end{figure}

Validations losses are plotted in \cref{fig:adapt-speedup}, on the left. Starting from an MLM-pretrained non-causal decoder model speeds up convergence significantly compared to training a causal-decoder model with an FLM objective from scratch.
To achieve a loss comparable to the one achieved after 168B tokens of FLM pretraining, language modeling adaptation requires only 105B additional tokens (a 1.6$\times$ speed-up).
This makes it possible to obtain both a high-quality zero-shot model and a good generative model, for only 1.6$\times$ the cost of training a single model. 

\paragraph{Non-causal masked language modeling adaptation (NC-A).} To investigate alternative avenues for adaptation, we now introduce \emph{non-causal masked language modeling adaptation}: starting from a causal decoder model pretrained with FLM as the objective, we then continue training the model as a non-causal decoder using an MLM objective. This is essentially the reverse of the language modeling adaptation setup, and the conversion is as easily undertaken by switching the attention mask.

Validation losses are plotted in \cref{fig:adapt-speedup}, on the right. Convergence on the MLM pretraining objective is significantly accelerated: by a factor of 3.3$\times$ compared to training a non-causal decoder from scratch, and up to a factor 9.1$\times$ compared to training a causal decoder from scratch (both with a masked language modeling objective). This is a significant improvement over even the previously considered language modeling adaptation, enabling one to obtain both a zero-shot model and an excellent generative model for only 1.3$\times$ the cost of training a single model. 

\begin{figure}[t]
    \centering
    \includegraphics[width=0.9\textwidth]{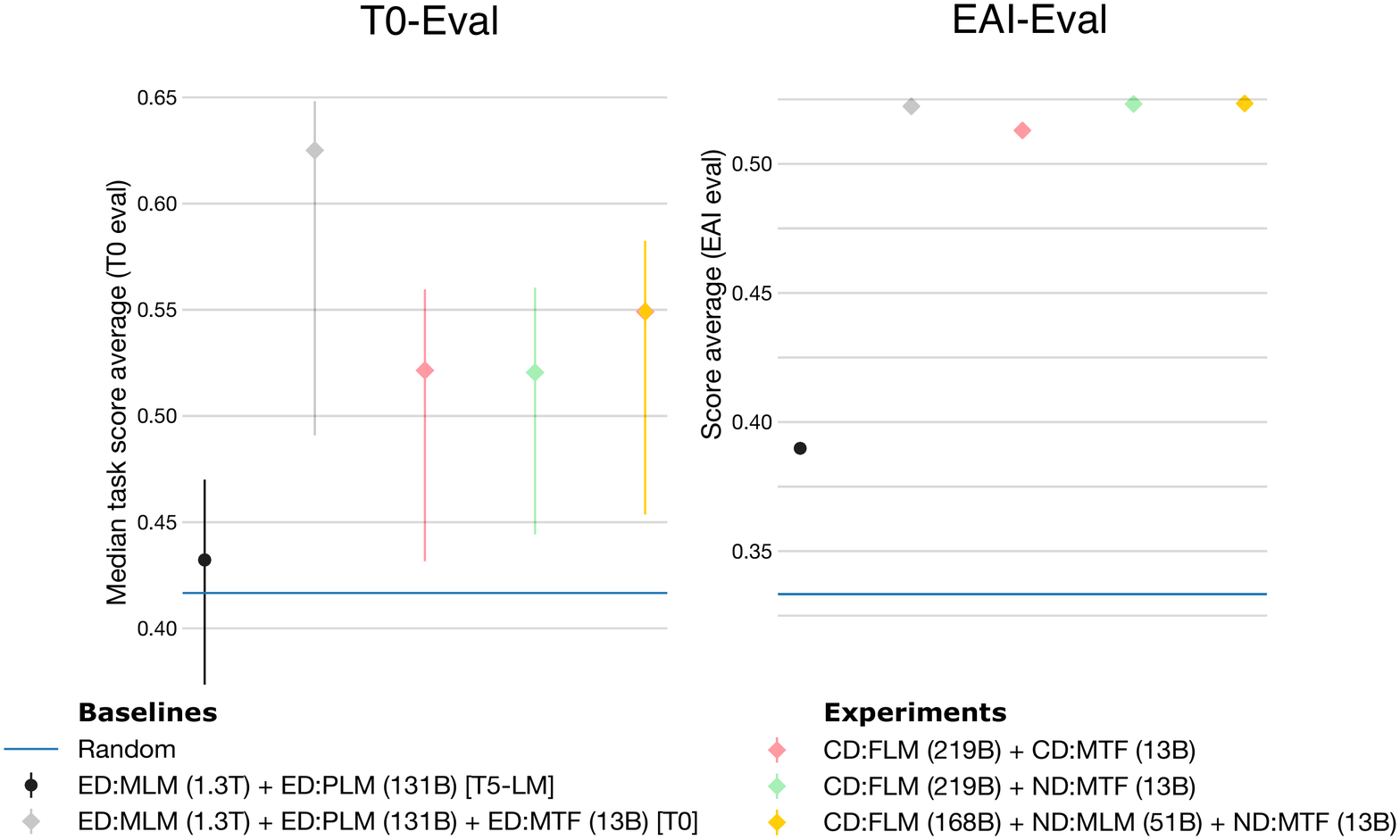}
    \caption{\textbf{Applying non-causal MLM adaptation to a causal decoder-only FLM before multitask finetuning improves zero-shot performance, even when controlling for additional LM pretraining for the same number of tokens.} Zero-shot generalization on T0-Eval (left) and EAI-Eval (right), for the T5-LM and T0 baselines (grey), and for models from our study. Converting the model into a non-causal decoder for multitask finetuning only does not improve performance on T0-Eval. Results after adaptation are in line with non-causal decoder-only pretrained with MLM in \cref{fig:flagship_models}.}
    \label{fig:adapt-fewshot-performance-improvement}
\end{figure}

Finally, we confirm that the improvement in validation loss also transfer to an improvement in zero-shot generalization. We evaluate the non-causal MLM adapted model, and check that it is better than the original causal decoder model pretrained with full language modeling, and control for the total number of training tokens.
Specifically, we evaluate zero-shot performance after multitask finetuning in three settings:
first, a causal decoder model pretrained with FLM for 219 billion tokens before being multitask finetuned; second, a causal decoder model pretrained with FLM for 219 billion tokens and then multitask finetuned as a non-causal decoder model; and, third, a causal decoder model first trained with FLM for 168 billion tokens, then MLM-adapted as an non-causal model for 51 billion tokens, and finally multitask finetuned.
All three variants are multitask finetuned for 13 billion tokens.
Results are presented \cref{fig:adapt-fewshot-performance-improvement}.
We find that the MLM-adapted model performs best by a significant margin and outperforms every other model we considered on EAI-Eval. Furthermore, the measured zero-shot generalization is in line with the MLM-pretrained non-causal decoder reported in \cref{fig:flagship_models}, though it still lags behind the MLM-pretrained encoder-decoder, despite the adapted models having seen 51 billion additional tokens.
Finally, we note that performing non-causal multitask finetuning of the causal model produces no meaningful change in performance.

\begin{mdframed}
\textbf{Finding 3.} Decoder-only models can be efficiently adapted from one architecture/objective prior to the other. Specifically, to obtain both a generative and a multitask model with the smallest total compute budget possible, we recommend starting with a \architecture{causal decoder-only} model, pretraining it with a \objective{full language modeling} objective, and then using \adaptation{non-causal masked language modeling adaptation} before taking it through \finetuning{multitask finetuning}.
\end{mdframed}


\section{Conclusion}
\label{sec:conclusion}
In this paper, we systematically studied the effects of pretraining objective and architecture choices on the zero-shot generalization abilities of large language models. Specifically, we compared language modeling and masked language modeling objectives applied to causal/non-causal decoder-only and encoder-decoder architectures.
We also evaluated zero-shot performance with and without multitask finetuning.
Notably, we found that the best objective and architecture is the \textit{opposite} in these two settings: a causal decoder-only pretrained with full language modeling performs best if evaluated immediately after pretraining, whereas when adding a multitask finetuning step, an encoder-decoder pretrained with masked language modeling performs best. We therefore evaluate the practice of adaptation, to convert models across architectures and objectives.
We found a simple efficient compromise, where a causal decoder-only model pretrained with full language modeling underwent additional masked language model training as a non-causal decoder-only model, yielding significant speedup in convergence over starting from scratch. This enables practitioners to get both an excellent generative model and a model that delivers good performance after multitask finetuning.
Our results provide significant new insights into the design of LLMs.
In the future, we are interested in work investigating architectures and objectives that perform well regardless of whether multitask finetuning is performed.
To facilitate future work, we release all models, code, and data used in our study.

\clearpage

\section*{Acknowledgements}
\label{sec:sup-ack}
\paragraph{BigScience.} This work was pursued as part of the Big Science research workshop, a one-year long initiative on large multilingual models and datasets. Specifically, this work was conducted by a task force within the architecture \& scaling group, seeking to establish the optimal architecture and pretraining objective for the final 176B parameter model produced by BigScience. We would also like to thank Stella Biderman for valuable comments.

\paragraph{Compute.} We thank the TPU Research Cloud team for providing us with generous access to TPUv4. We thank the TPUv4 Alpha team for providing technical support for this work, and notably James Bradbury. All pretraining, adaptations, finetuning, and evaluations featured in this paper used TPUv4. 

This work was granted access to the HPC resources of Institut du développement et des ressources en informatique scientifique (IDRIS) du Centre national de la recherche scientifique (CNRS) under the allocation 2021-A0101012475 made by Grand équipement national de calcul intensif (GENCI). Specifically, early experiments on non-causal decoder models, which sparked this exhaustive study, were performed on the Jean Zay supercomputer of IDRIS. 

\paragraph{Author-specific funding.}
\vspace{0.5cm}
\begin{center}
\includegraphics[width = 0.12\textwidth]{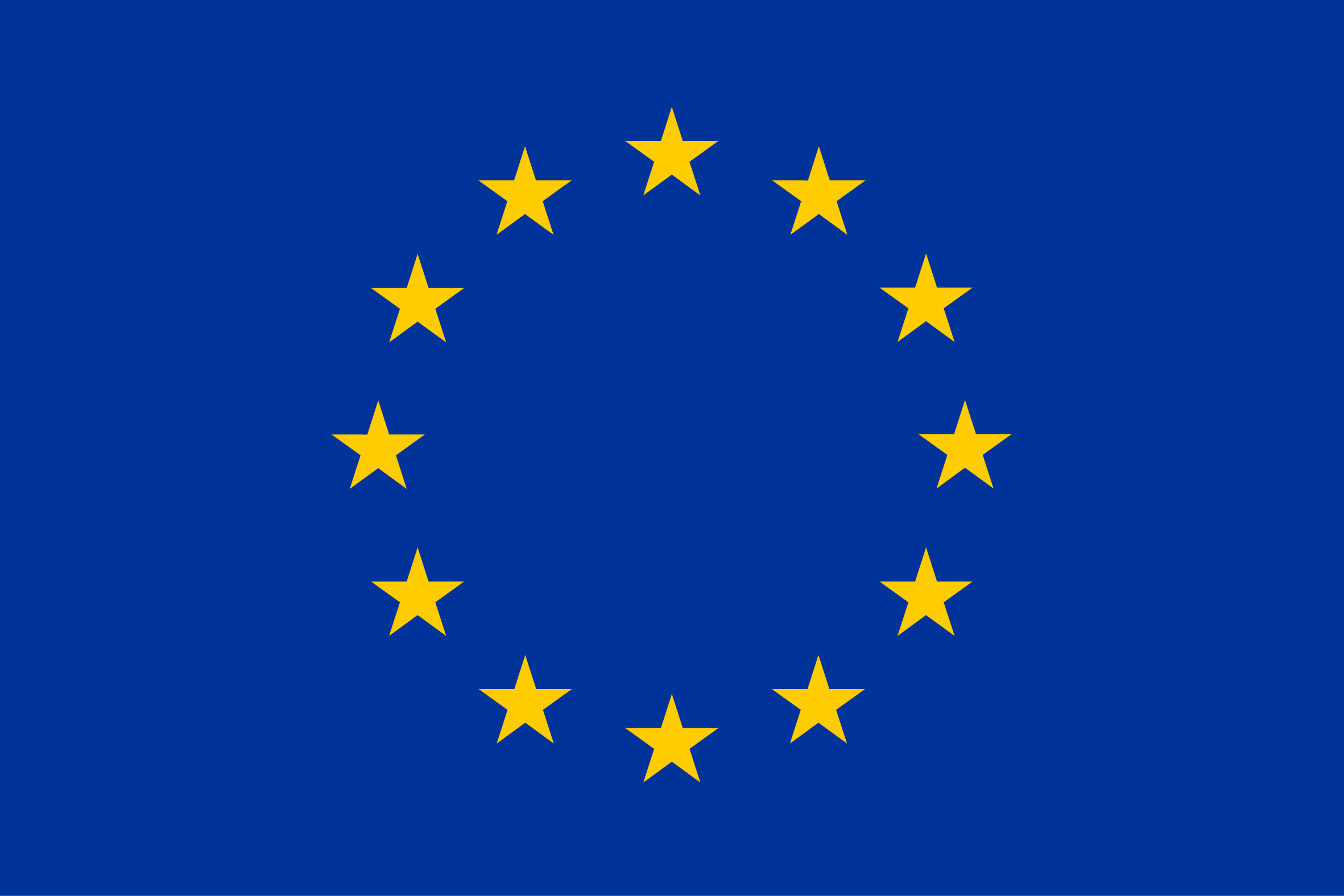}\\
Daniel Hesslow has received funding from the European Union’s Horizon 2020 research and innovation programme under the Marie Skłodowska-Curie grant agreement No 860360.
\end{center}

\bibliography{references}
\bibliographystyle{unsrtnat}

\newpage
\appendix

\section{Contributions}
\label{sec:sup-contributions}
Thomas Wang wrote code, ran experiments, performed evaluation, generated plots, and helped with paper writing. Adam Roberts led the creation of the codebase used in this project, proposed some experiments, ran all of the final experiments, generated plots, and helped with paper writing. Daniel Hesslow made it possible to evaluate models with the EleutherAI harness, created diagrams, and helped with paper writing. Teven Le Scao ran evaluations and plotted results. Hyung Won Chung implemented infrastructure code for different architectural and objective variants. Iz Beltagy co-chaired the BigScience Architecture \& Scaling Group and helped with paper editing. Julien Launay co-chaired the BigScience Architecture \& Scaling Group and had the largest role in paper writing. Colin Raffel proposed the project, the experiments, and the adaptation methods and wrote portions of the paper.

\section{Broader impacts}
\subsection{Societal impact}
\label{sec:sup-ethics}
The risks and societal challenges raised by large language models have been discussed extensively in the literature \citep{solaiman2019release, bommasani2021opportunities}. Our research is strictly oriented on benchmarking modeling aspects, and thus does not introduce any novel challenge beside those already identified. Notably, many similarly capable models have already been released publicly in the past \citep{T5, gpt-j, T0, blackgpt}.

In the spirit of reproducibility and openness, we release all artefacts produced during this study: the configs necessary to reproduce our results from scratch, checkpoints of all of the models trained, and detailed evaluation results. These artefacts are intended for research only: we did not evaluate the potential biases of the models trained, and cannot guarantee they won't produce harmful content. Accordingly, \textbf{these models should not be used in production or exposed to the public.}

We also highlight that algorithmic choices can introduce biases on their own \citep{hooker2021moving}: one limitation of our study is that we did not explore whether specific architectures and objectives had an impact on the toxicity and biases of a given model. However, the public availability of all the models trained for this study enables researchers to conduct such a follow-up study at minimal compute cost. 

\subsection{Environmental impact}
\label{sec:sup-carbon}
Across all experiments undertaken over the course of this study (including unreported preliminary and failed experiments), we performed training for 1,854 hours on 64 chips (TPUv4-128) and 1,395 hours on 512 TPUv4 chips (TPUv4-1024) for a total of 832,896 chip-hours. Recently, \citet{palm} presented the results of training a 540 billion parameter language model on TPUv4 chips in the same datacenter where we ran our experiments. Their model was trained for 1,200 hours on 6,144 TPUv4 chips and 336 hours on 3,072 TPUv4 chips, for a total of 8,404,992 chip-hours. \citet{palm} estimates the carbon emissions of their model training to be 240.5 tCO2e based on the net tCO2e per MWh of the datacenter during training and the energy usage of TPUv4 chips. We therefore estimate our carbon emissions to be approximately 23.8 tCO2e, which is approximately half of what \cite{patterson2021carbon} report for the original T5 model training (46.7 tCO2).

\section{Implementation: prefix language modeling for encoder-decoder on TPU}
\label{sec:sup-tpu}
Due to constant size constraints, for encoder-decoder with prefix language modeling, we have to concatenate two examples of 626 tokens into one. We randomly sample an index $i$ between 0 and 626, and use $i$ and $626 - i$ as prefix indices in the two examples. We use masking to keep them independent throughout training. The encoder-decoder thus has a 1,252 sequence length, and we train it with a batch size of 1,024 sequences instead of 2,048 to keep the number of tokens constant.

\newpage

\section{Evaluation: benchmarks composition and baselines}
\label{sec:sup-eval}
We detail the split across EAI-Eval and T0-Eval in \cref{tab:eval-tasks}, and provide random baselines in \cref{tab:random-baselines}. 

\vfill

\begin{table}[h]
\caption{\textbf{Tasks used for zero-shot evaluation within EAI-Eval and T0-Eval, as well as tasks included in the T0 training set for multitask finetuning.} Note that T0-Eval/Training tasks include multiple prompts, while EAI-Eval tasks have a single prompt. Although tasks are shared between the two, there are no shared prompts between EAI-Eval and T0-Eval.}
\label{tab:eval-tasks}
\begin{center}
\begin{tiny}
\begin{tabular}{@{}lllccc@{}}
\toprule
\multicolumn{2}{c}{\sc{Task}}           & \multicolumn{1}{c}{\sc{Type}}           & \multicolumn{3}{c}{\sc{Dataset}}                                    \\ 
             &           &                                    & T0 Training          & EAI-Eval                                              & T0-Eval                                               \\\midrule
ANLI         &           & Natural Language Inference         &                           & \checkmark                     & \checkmark                     \\
ARC \citep{clark2018arc}           & Challenge & Closed-Book Question Answering     &                           & \checkmark & \multicolumn{1}{c}{}                          \\
             & Easy      & Closed-Book Question Answering     &                           & \checkmark & \multicolumn{1}{c}{}                          \\
GLUE         & MRPC   \citep{dolan2016mrpc}   & Paraphrase Identification          & \checkmark & \checkmark                     &                                               \\
             & QQP    \citep{iyer2019qqp}    & Paraphrase Identification          & \checkmark & \checkmark                     &                                               \\
HEAD-QA      &           & Multiple-Choice Question Answering &                           & \checkmark                     &                                               \\
HellaSwag  \citep{zellers2019hellaswag}  &           & Sentence Completion                &                           & \checkmark                     & \checkmark                     \\
LAMBADA     \citep{paperno2016lambada}   &           & Sentence Completion                &                           & \checkmark                     &                                               \\
LogiQA       \citep{liu2020logiqa}  &           & Multiple-Choice Question Answering &                           & \checkmark                     &                                               \\
MathQA   \citep{amini2019mathqa}      &           & Multiple-Choice Question Answering &                           & \checkmark                     &                                               \\
OpenBookQA   \citep{mihaylov2press2021train018openbookqa} &           & Multiple-Choice Question Answering &                           & \checkmark                     &                                               \\
PIQA     \citep{bisk2020piqa}      &           & Multiple-Choice Question Answering &                           & \checkmark                     &                                               \\
PROST  \citep{aroca-ouellette2021prost}        &           & Multiple-Choice Question Answering &                           & \checkmark                     &                                               \\
PudMedQA    \citep{jin2019pubmedqa}    &           & Multiple-Choice Question Answering &                           & \checkmark                     &                                               \\
QNLI   \citep{rajpurkar2016squad,wang2019glue}        &           & Sentence Completion                &                           & \checkmark                     &                                               \\
Race   \cite{lai2017large}       &           & Multiple-Choice Question Answering &                           & \checkmark & \multicolumn{1}{c}{}                          \\
SciQ  \citep{welbl2017sciq}        &           & Multiple-Choice Question Answering & \checkmark & \checkmark                     &                                               \\
SST    \citep{socher2013sst}       \citep{socher2013sst}  &           & Sentiment                          &                           & \checkmark                     &                                               \\
StoryCloze   &           & Sentence Completion                &                           &                                               & \checkmark                     \\
SuperGLUE    & Boolq  \citep{clark2019boolq}   & Multiple-Choice Question Answering &                           & \checkmark & \multicolumn{1}{c}{}                          \\
             & CB        & Natural Language Inference         &                           & \checkmark & \checkmark \\
             & COPA  \citep{gordon2012copa}    & Sentence Completion                &                           & \checkmark & \checkmark \\
             & MultiRC \citep{kashabi2018multirc}  & Multiple-Choice Question Answering &                           & \checkmark                     &                                               \\
             & RTE   \citep{dagan2005rte}     & Natural Language Inference         &                           & \checkmark                     & \checkmark                     \\
             & WIC   \citep{pilehavar2018wic}    & Word Sense Disambiguation          &                           & \checkmark                     & \checkmark                     \\
             & WSC \citep{levesque2012winograd}  & Word Sense Disambiguation          &                           & \checkmark                     & \checkmark                     \\
TriviaQA     \citep{joshi2017triviaqa}  &           & Closed-Book Question Answering     &                           & \checkmark & \multicolumn{1}{c}{}                          \\
WebQuestions \citep{berant2013semantic} &           & Closed-Book Question Answering     &                           & \checkmark                     &                                               \\
Winogrande  \citep{sakaguchi2019winogrande}    &           & Coreference resolution             &                           & \checkmark                     & \checkmark                     \\
WNLI  \citep{sakaguchi2019winogrande}          &           & Natural language inference         &                           & \checkmark                     &                           \\ \bottomrule                   
\end{tabular}
\end{tiny}
\end{center}
\vskip -0.1in
\end{table}

\vfill

\newpage
\null
\vfill
\begin{table}[h]
\caption{\textbf{Random baselines for all tasks considered across EAI harness and T0 Eval.} These baselines were obtained from the papers introducing these tasks.}
\label{tab:random-baselines}
\vskip 0.15in
\begin{center}
\begin{small}
\begin{tabular}{@{}llc@{}}
\toprule
\multicolumn{2}{c}{\sc{Task}}           & \sc{Random baseline}                                \\ 
\midrule
ANLI         &           & 33.3  \\
ARC          & Challenge & 25.0       \\
             & Easy      & 25.0                         \\
GLUE         & MRPC      & 50.0                                                  \\
             & QQP       & 50.0                                                             \\
HEAD-QA      &           & 25.0                                 \\
HellaSwag    &           & 25.0                 \\
LAMBADA      &           & 0.0                                        \\
LogiQA       &           & 25.0                                               \\
MathQA       &           & 20.1                                          \\
OpenBookQA   &           & 25.0       \\
PIQA         &           & 50.0        \\
PROST        &           & 25.0                          \\
PudMedQA     &           & 33.3                                         \\
QNLI         &           & 50.0                                         \\
Race         &           & 25.0                       \\
SciQ         &           & 25.0                                          \\
SST          &           & 50.0                                                \\
StoryCloze   &           & 50.0                 \\
SuperGLUE    & Boolq     & 50.0                     \\
             & CB        & 50.0 \\
             & COPA      & 50.0 \\
             & MultiRC   & 5.8                                          \\
             & RTE       & 50.0                     \\
             & WIC       & 50.0                     \\
             & WSC       & 50.0                     \\
TriviaQA     &           & 0.0                      \\
WebQuestions &           & 0.0                        \\
Winogrande   &           & 50.0             \\
WNLI         &           & 50.0      \\ \midrule
\sc{EAI-Eval} & & 33.3 \\
\sc{T0-Eval} & & 41.7 \\\bottomrule                   
\end{tabular}
\end{small}
\end{center}
\vskip -0.1in
\end{table}
\vfill
\newpage
\FloatBarrier

\section{Additional results}
\subsection{Preliminary results and evolution throughout pre-training}
\label{sec:sup-res}
Leveraging early pretraining results at 42B and 84B tokens, we motivate in this section two special design decisions in our study:
\begin{itemize}
    \item \textbf{Not considering span corruption for evaluation after pretraining only.} In \cref{tab:lm-results}, we only report zero-shot generalization results immediately after pretraining for the full and prefix language modeling objectives. We choose not to report results when using a masked language modeling objective, as \cref{tab:eai-pre-62} demonstrates that after 84B tokens of pretraining, models pretrained with this objective still achieve close to random performance, and severely underperform models pretrained with prefix or full language modeling.
    \item \textbf{Not systematically performing LM adaptation before multitask finetuning.} \citet{T0} originally perform LM adaptation before multitask finetuning. As outlined in \cref{tab:lm-plus-t0}, using early models pretrained for 42B tokens, we found this practice did not consistently improve zero-shot generalization, and in fact worsened it in most cases. Accordingly, results in \cref{fig:flagship_models} do not use LM adaptation before multitask finetuning. This is in line with the findings of \citet{lester2021power} that larger models (of the same scale that we are considering in our study) do not benefit from performing LM adaptation before prompt tuning.
\end{itemize}

\begin{table}[h]
\caption{\textbf{Models pretrained with masked language modeling achieve performance close to the random 33.3$\%$ baseline on EAI-Eval, significantly underperforming full and prefix language modeling.} Average accuracy on EAI-Eval after pretraining for 84B tokens. This observations leads us to not consider masked language modeling for evaluations after pretraining only.}
\label{tab:eai-pre-62}
\vskip 0.15in
\begin{center}
\begin{small}
\begin{tabular}{@{}lc@{}}
\toprule
                   & \sc{EAI-Eval} \\ \midrule
\sc{FLM/PLM}  &            \\
Causal decoder     & \cellcolor{neutral}\textbf{42.4}                           \\
Non-causal decoder & 42.2                            \\
Encoder-decoder    & 39.6                             \\ \midrule
\sc{MLM}    &                                   \\
Causal decoder     & 37.8                              \\
Non-causal decoder & 37.7                          \\
Encoder-decoder    & 34.6                     \\ \bottomrule
\end{tabular}
\end{small}
\end{center}
\vskip -0.1in
\end{table}

\begin{table}[h]
\caption{\textbf{Performing LM adaptation before multitask finetuning does not improve results, and in fact hinders performance in most cases.} Average accuracy on EAI-Eval and T0-Eval for different adaptation strategies after 42B tokens of masked language modeling pretraining. LM adaptation alone is insufficient, and most performance gains come from MT finetuning. Accordingly, we diverge from the setup of \citet{T0}, and forego systematic LM adaptation before multitask finetuning.}
\label{tab:lm-plus-t0}
\vskip 0.15in
\begin{center}
\begin{small}
\begin{tabular}{@{}lcc@{}}
\toprule
                   & \sc{EAI-Eval} & \sc{T0-Eval}  \\ \midrule
\sc{LM adaptation}  &            \\
Causal decoder     & 38.6 & 43.9                          \\
Non-causal decoder & 39.5 & 40.8                            \\
Encoder-decoder    & 38.6 & 39.1                            \\ \midrule
\sc{MT finetuning}    &                                   \\
Causal decoder     & 43.3 & 45.8                              \\
Non-causal decoder & \cellcolor{finetuning}\textbf{45.9} & 48.9                         \\
Encoder-decoder    & 45.4 & \cellcolor{finetuning}\textbf{53.7}                     \\ \bottomrule
\sc{LM-A+MT finetuning}    &                                   \\
Causal decoder     & 43.9 & 46.7                              \\
Non-causal decoder & 45.0 & 48.0                          \\
Encoder-decoder    & 45.7 & 52.6                     \\ \bottomrule
\end{tabular}
\end{small}
\end{center}
\vskip -0.1in
\end{table}

\newpage

\subsection{Complete Results}
\label{sec:sup_complete-results}

\begin{figure*}[h]
    \centering
    \includegraphics[width=\textwidth]{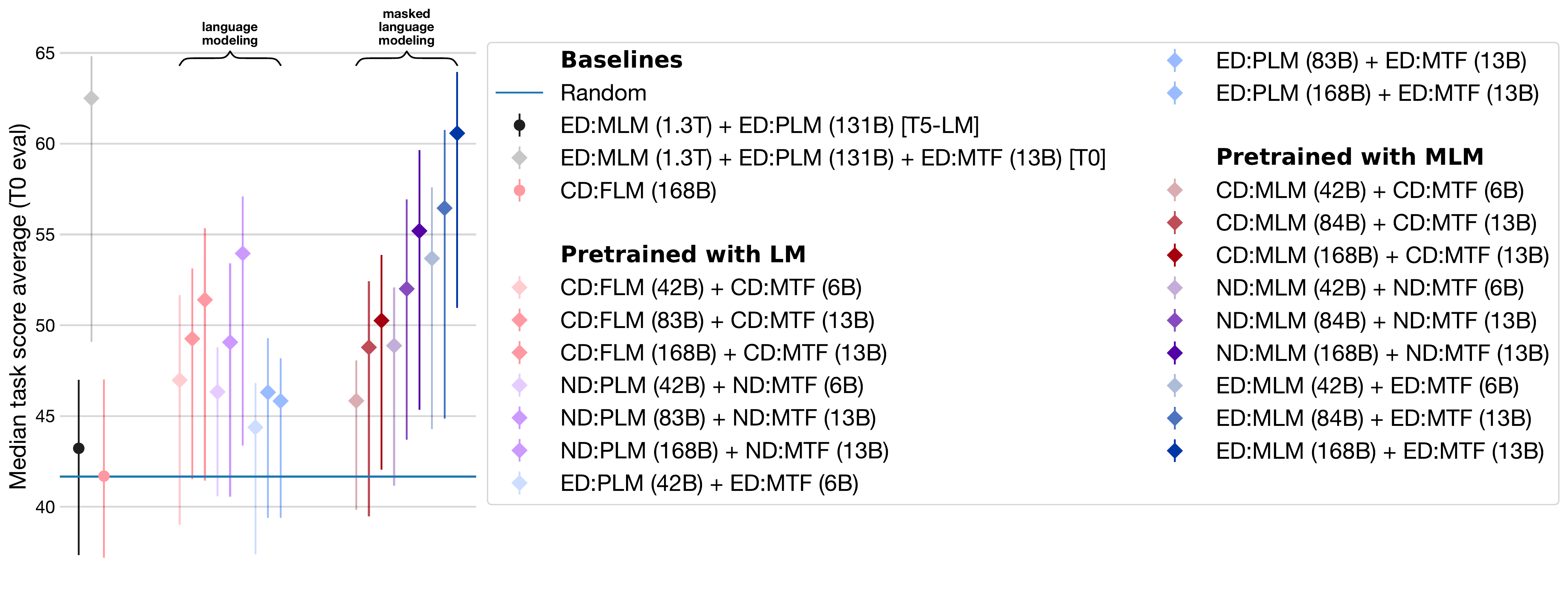}
    \caption{\textbf{Performance on T0-Eval after multitask finetuning for increasing amounts of pretraining (measured in tokens).} Our best model, an encoder-decoder trained with masked language modeling, is already above the final performance of the other configurations with only a quarter of the pretraining tokens. Note that the ordering does not change significantly throughout pretraining.}
    \label{fig:through_training}
\end{figure*}

We report results for all intermediary checkpoints produced in \cref{tab:all-exps}, and specifically for all multitask finetuned checkpoints on T0-Eval in \cref{fig:through_training}.

\vfill

\begin{table}[h]
\caption{\textbf{Average accuracy on EAI-EVal and T0 Eval for all experiments.} Experiments are represented as a combination of \textit{architecture}:\textit{objective} (\textit{tokens}) training stages, where \textit{architecture} is one of causal decoder-only (\texttt{CD}), non-causal decoder-only (\texttt{ND}), or encoder-decoder (\texttt{ED}), and \textit{objective} is one of full language modeling (\texttt{FLM}), prefix language modeling (\texttt{PLM}), masked language modeling (\texttt{MLM}), or multitask finetuning (\texttt{MTF}).}
\label{tab:all-exps}
\vskip 0.15in
\begin{center}
\begin{small}
\begin{tabular}{@{}lllccc@{}}
\toprule
\multicolumn{3}{c}{\sc{Training Stage}} &\sc{Total Tokens}&\sc{EAI-Eval}&\sc{T0-Eval} \\
\sc{Pretraining}&\sc{Adaptation}&\sc{Finetuning} &&&  \\ \midrule
\texttt{CD:MLM (38B)} & \texttt{CD:FLM (4B)}     && 42B & 38.6 & 43.9                         \\
\texttt{ND:MLM (38B)} & \texttt{ND:PLM (4B)}     && 42B & 39.5 & 40.8                         \\
\texttt{ED:MLM (38B)} & \texttt{ED:MLM (4B)}     && 42B & 38.6 & 39.1                         \\     
\texttt{CD:MLM (42B)} && \texttt{CD:MTF (13B)}   & 55B & 43.3 & 45.8                         \\
\texttt{ND:MLM (42B)} && \texttt{ND:MTF (13B)}   & 55B & 45.9 & 48.9                         \\
\texttt{ED:MLM (42B)} && \texttt{ED:MTF (13B)}   & 55B & 45.4 & 53.7                         \\
\texttt{CD:MLM (38B)} & \texttt{CD:FLM (4B)} & \texttt{CD:MTF (13B)} & 55B & 43.9 & 46.7                \\
\texttt{ND:MLM (38B)} & \texttt{ND:PLM (4B)} & \texttt{ND:MTF (13B)} & 55B & 45.0 & 48.0                 \\
\texttt{ED:MLM (38B)} & \texttt{ED:PLM (4B)} & \texttt{ED:MTF (13B)} & 55B & 45.7 & 52.6           \\
\texttt{CD:FLM (84B)}                 &&& 84B & 42.4 & -                    \\
\texttt{ND:PLM (84B)}                 &&& 84B & 42.2 & -                            \\
\texttt{ED:PLM (84B)}                 &&& 84B & 39.6 & -                            \\
\texttt{CD:MLM (84B)}                 &&& 84B & 37.8 & -                            \\
\texttt{ND:MLM (84B)}                 &&& 84B & 37.7 & -                            \\
\texttt{ED:MLM (84B)}                 &&& 84B & 34.6 & -                            \\
\texttt{CD:FLM (84B)} && \texttt{CD:MTF (13B)}   & 97B & 49.0 & 49.9                        \\
\texttt{ND:PLM (84B)} && \texttt{ND:MTF (13B)}   & 97B & 46.3 & 50.0                         \\
\texttt{ED:PLM (84B)} && \texttt{ED:MTF (13B)}   & 97B & 43.2 & 46.5                         \\
\texttt{CD:MLM (84B)} && \texttt{CD:MTF (13B)}   & 97B & 45.8 & 48.2                         \\
\texttt{ND:MLM (84B)} && \texttt{ND:MTF (13B)}   & 97B & 49.0 & 52.6                         \\
\texttt{ED:MLM (84B)} && \texttt{ED:MTF (13B)}   & 97B & 49.0 & 56.5                         \\
\texttt{CD:FLM (168B)}                &&& 168B & 44.2 & 42.4                        \\
\texttt{ND:PLM (168B)}                &&& 168B & 43.5 & 41.8                        \\
\texttt{ED:PLM (168B)}                &&& 168B & 39.9 & 41.7                        \\
\texttt{CD:FLM (168B)} && \texttt{CD:MTF (13B)}  & 181B & 50.4 & 51.4                        \\
\texttt{ND:PLM (168B)} && \texttt{ND:MTF (13B)}  & 181B & 48.9 & 54.0                        \\
\texttt{ED:PLM (168B)} && \texttt{ED:MTF (13B)}  & 181B & 44.2 & 45.8                        \\
\texttt{CD:MLM (168B)} && \texttt{CD:MTF (13B)}  & 181B & 47.1 & 50.3                        \\
\texttt{ND:MLM (168B)} && \texttt{ND:MTF (13B)}  & 181B & 51.0 & 55.2                        \\
\texttt{ED:MLM (168B)} && \texttt{ED:MTF (13B)}  & 181B & 51.3 & 60.6                        \\
\texttt{CD:FLM (168B)} & \texttt{CD:FLM (51B)} & \texttt{CD:MTF (13B)}  & 232B & 51.3 & 52.1          \\       
\texttt{CD:FLM (168B)} & \texttt{CD:FLM (51B)} & \texttt{ND:MTF (13B)}  & 232B & 52.3 & 52.0          \\
\texttt{CD:FLM (168B)} & \texttt{ND:MLM (51B)} & \texttt{ND:MTF (13B)}  & 232B & 52.3 & 54.9          \\
\midrule
\multicolumn{3}{@{} l}{\sc{T5-LM Baseline} \citep{lester2021power}} \\
\texttt{ED:MLM (1.28T)} & \texttt{ED:PLM (131B)} && 1.41T & 39.0 & 43.2 \\
\midrule
\multicolumn{3}{@{} l}{\sc{T0 Baseline} \citep{T0}} \\
\texttt{ED:MLM (1.28T)} & \texttt{ED:PLM (131B)} & \texttt{ED:MTF (13B)} & 1.43T & 52.2 & 62.5 \\
\midrule
\multicolumn{3}{@{} l}{\sc{Random Baseline}} & - & 32.9 & 41.7 \\
\bottomrule
\end{tabular}
\end{small}
\end{center}
\vskip -0.1in
\end{table}

\vfill
\newpage

\subsection{Impact of dropout on multitask finetuning}
\label{sec:sup-dropout}
\begin{figure}[h]
    \centering
    \includegraphics[width=0.9\textwidth]{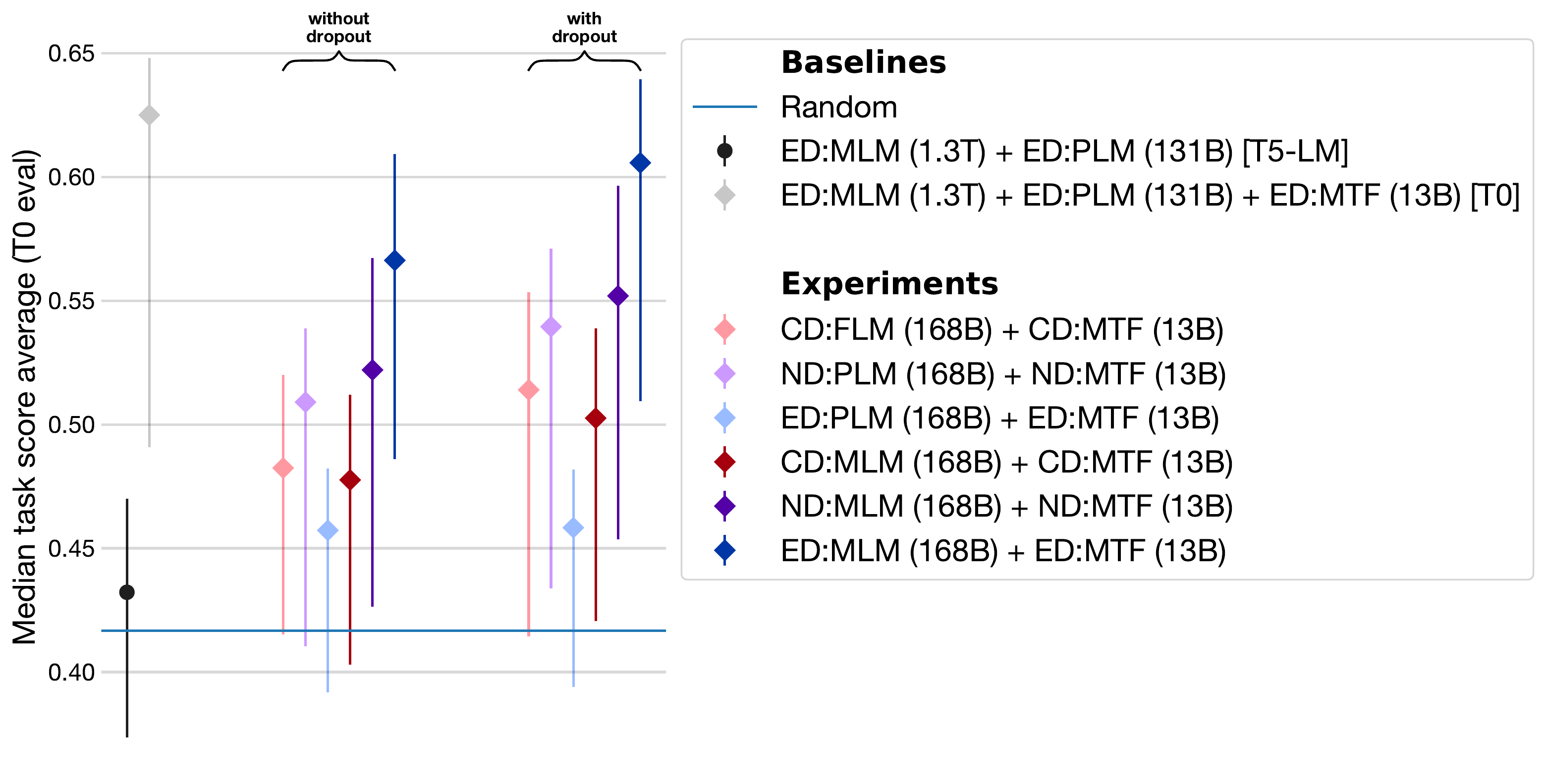}
    \caption{\textbf{Using dropout during multitask finetuning improves zero-shot generalization.} Performance on T0-Eval with and without dropout. The impact of dropout is proportionally similar across architecture and objectives, not benefitting any specific configuration more.}
    \label{fig:dropout}
\end{figure}

We also performed multitask finetuning without using dropout, with results in \cref{fig:dropout}. We find that using dropout as originally suggested by \citet{T0} significantly boosts zero-shot generalization. Results are consistent across architectures and pretraining objectives.

\newpage
\subsection{Adaptation from an encoder-decoder}
\label{sec:sup-ec-adapt}
\begin{figure}[h]
    \centering
    \includegraphics[width=0.7\textwidth]{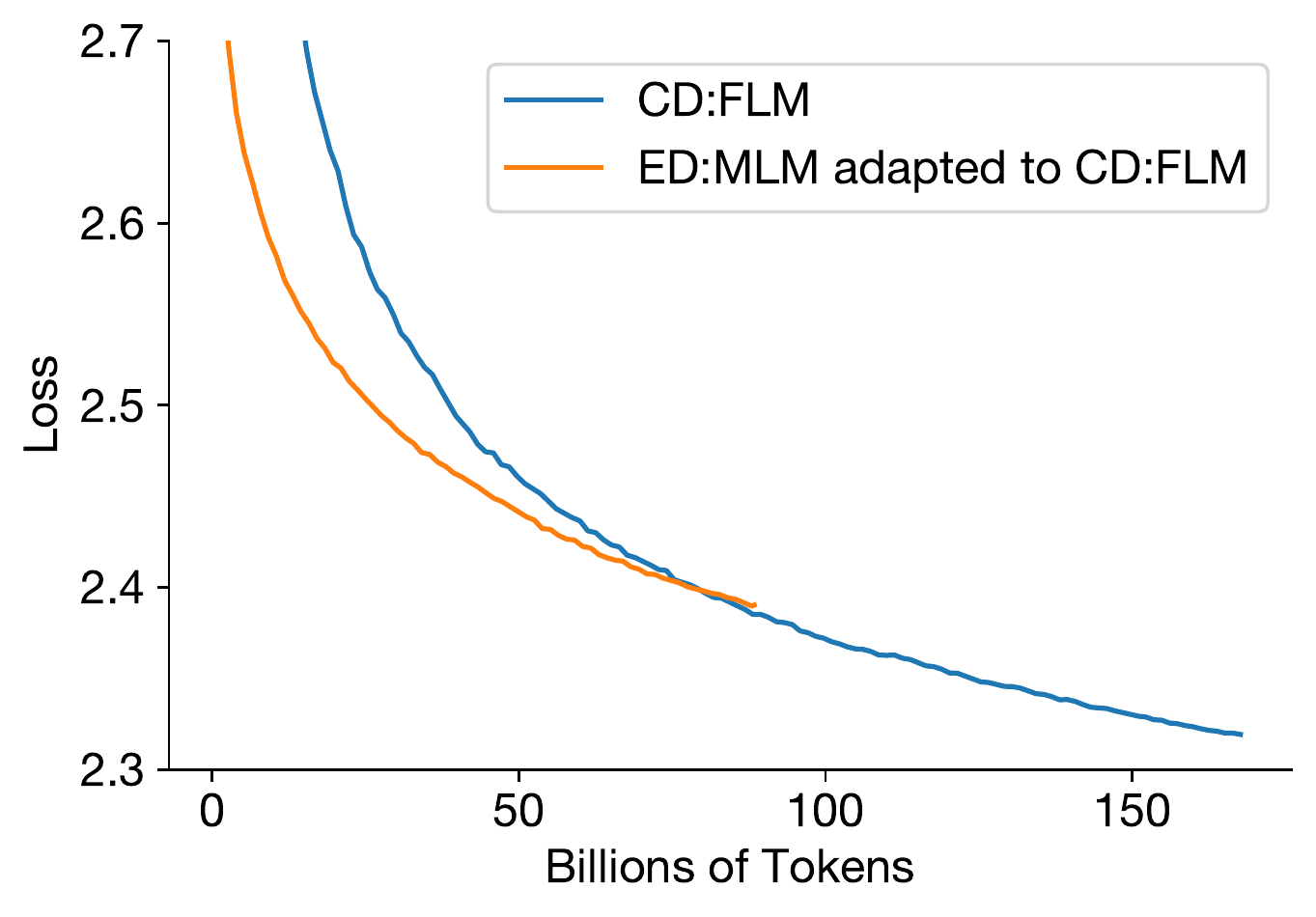}
    \caption{\textbf{Converting an encoder-decoder pretrained with MLM to a causal decoder-only using FLM leads to worse performance compared to training from scratch.} Validation loss when adapting an encoder-decoder pretrained with MLM to a causal decoder-only using FLM. We adapted a pretrained (for 168B tokens) encoder-decoder model to decoder-only by feeding an empty input into the encoder and causally training with a FLM objective on the decoder. We stopped this adaptation once it was clear the performance would not match that of a causal FLM trained from scratch, in contrast with the other adaptations we studied.}
    \label{fig:ec-cd-adapt}
\end{figure}

When studying adaptation and the conversion from one architecture to another, we also considered converting to and from encoder-decoder models. Conversion across causal and non-causal decoder-only models is straightforward, simply by switching the attention mask; for encoder-decoder, parameters have to be either pruned or added for both the entire encoder, and for the cross-attention in the decoder. Results from one of our attempt to convert an encoder-decoder into a causal decoder are reported in \cref{fig:ec-cd-adapt}. While converting across causal/non-causal decoder provides an improvement over training from scratch, this is not the case here.

\end{document}